\documentclass[sigconf]{acmart}

\AtBeginDocument{%
  }
    
\setcopyright{acmlicensed}

\copyrightyear{2025}
\acmYear{2025}
\setcopyright{cc}
\setcctype{by}
\acmConference[KDD '25]{Proceedings of the 31st ACM SIGKDD Conference on
Knowledge Discovery and Data Mining V.2}{August 3--7, 2025}{Toronto, ON,
Canada}
\acmBooktitle{Proceedings of the 31st ACM SIGKDD Conference on Knowledge
Discovery and Data Mining V.2 (KDD '25), August 3--7, 2025, Toronto, ON,
Canada}
\acmDOI{10.1145/3711896.3736559}
\acmISBN{979-8-4007-1454-2/2025/08}

\usepackage{paralist,xcolor}
\usepackage{bm}
\usepackage{multirow}
\usepackage[export]{adjustbox}

\usepackage{colortbl}
\definecolor{lightgray}{gray}{.9}
\definecolor{deepgray}{gray}{.8}

\usepackage{threeparttable}
\newcolumntype{I}{!{\vrule width 1pt}}
\makeatletter
\newcommand{\thickhline}{%
    \noalign {\ifnum 0=`}\fi \hrule height 1pt
    \futurelet \reserved@a \@xhline
}
\makeatother
\definecolor{mygray}{gray}{.9}
\usepackage{float}
\usepackage[ruled, vlined]{algorithm2e}
\usepackage{pifont}
\usepackage{ragged2e}
\usepackage{paralist}
\usepackage[toc,page]{appendix}
\usepackage{url}

\definecolor{mygray}{gray}{.9}
\definecolor{mygreen}{RGB}{93,173,85}
\definecolor{mywarning}{RGB}{233,144,61}
\definecolor{DarkRed}{RGB}{0,0,0}
\definecolor{azure}{rgb}{0.0, 0.5, 1.0}
\definecolor{gray}{rgb}{0.3, 0.3, 0.3}
\definecolor{DarkGreen}{RGB}{42,110,63}
\definecolor{linkblue}{RGB}{0, 0, 255} 

\usepackage[capitalize]{cleveref}
\crefname{section}{Sec.}{Secs.}
\crefname{table}{Tab.}{Tabs.}
\crefname{section}{§}{§§}
\usepackage[toc,page]{appendix}
\usepackage{hyperref}
\usepackage{xcolor}
\hypersetup{
    colorlinks=true,
    urlcolor=blue, 
}

\newcounter{bulletsec}
\renewcommand{\thebulletsec}{\arabic{bulletsec}}

\crefname{bulletsec}{§}{§§}
\Crefname{bulletsec}{§}{§§}

\makeatletter
\DeclareRobustCommand\onedot{\futurelet\@let@token\@onedot}
\def\@onedot{\ifx\@let@token.\else.\null\fi\xspace}

\usepackage{amsmath}
\usepackage{mathtools}
\usepackage{amsthm}

\begin{document}


\title{Graph ODEs and Beyond: A Comprehensive Survey on Integrating Differential Equations with Graph Neural Networks}

\author{Zewen Liu}
\authornote{Both authors contributed equally to this research.}
\affiliation{%
\department{Emory University}
  \country{}
}
\email{zewen.liu@emory.edu}

\author{Xiaoda Wang}
\authornotemark[1]
\affiliation{%
\department{Emory University}
  \country{}
}
\email{xiaoda.wang@emory.edu}

\author{Bohan Wang}
\affiliation{%
    \department{Emory University}
  \country{}
}
\email{bohan.wang2@emory.edu}

\author{Zijie Huang}
\affiliation{%
\department{Amazon}
  \country{}
}
\email{zijiehjj@gmail.com}

\author{Carl Yang}
\affiliation{%
\department{Emory University}
  \country{}
}
\email{j.carlyang@emory.edu}

\author{Wei Jin}

\affiliation{%
\department{Emory University}
  \country{}
}
\email{wei.jin@emory.edu}

\begin{abstract}
Graph Neural Networks (GNNs) and differential equations (DEs) are two rapidly advancing areas of research that have shown remarkable synergy in recent years. GNNs have emerged as powerful tools for learning on graph-structured data, while differential equations provide a principled framework for modeling continuous dynamics across time and space. The intersection of these fields has led to innovative approaches that leverage the strengths of both, enabling applications in physics-informed learning, spatiotemporal modeling, and scientific computing. This survey aims to provide a comprehensive overview of the burgeoning research at the intersection of GNNs and DEs. We will categorize existing methods, discuss their underlying principles, and highlight their applications across domains such as molecular modeling, traffic prediction, and epidemic spreading. Furthermore, we identify open challenges and outline future research directions to advance this interdisciplinary field. A comprehensive paper list is provided at \href{https://github.com/Emory-Melody/Awesome-Graph-NDEs}{\textcolor{blue}{https://github.com/Emory-Melody/Awesome-Graph-NDEs}}. This survey serves as a resource for researchers and practitioners seeking to understand and contribute to the fusion of GNNs and DEs.

\end{abstract}



\maketitle

\section{Introduction}
Understanding and predicting complex behaviors in natural and engineered systems is a fundamental challenge across scientific and industrial domains. Many real-world phenomena exhibit dynamic evolution over time, governed by intricate interdependencies between variables. Examples include climate patterns shaped by atmospheric and oceanic interactions~\cite{keane2017climate}, population dynamics influenced by birth and migration rates~\cite{holmes1994partial}, financial markets driven by investor behavior and economic indicators~\cite{yang2020parameter}, disease progression driven by biological factors~\cite{dang2023conditional,wang2025ncode}, and the spread of infectious diseases determined by transmission dynamics and intervention strategies~\cite{maki2013infectious, liu2024review}. Capturing these temporal changes and underlying mechanisms requires mathematical models that not only describe system behavior but also provide predictive insights.


To effectively model dynamical systems, Differential Equations (DEs), such as Ordinary Differential Equations (ODEs)~\cite{butcher2000numerical}, Partial Differential Equations (PDEs)~\cite{sloan2012partial}, and Stochastic Differential Equations (SDEs)~\cite{van1976stochastic}, relate one or more unknown functions to their derivatives, thus describing how outputs vary given changing variables. At their core, DEs consist of three essential components: (1) \textit{state variables} that describe the system’s condition, 
(2) \textit{derivatives} that model and capture the rate of change, and (3) \textit{parameters} that influence the dynamics under given initial and boundary conditions. These elements work together to provide a structured approach to understanding how systems evolve over time.

Despite the crucial role of DEs in modeling complex phenomena, various challenges arise from real-world applications. Notably, many systems exhibit intricate, high-dimensional dynamics that are difficult to capture using purely knowledge-driven DE formulations~\cite{todorovski2007integrating}, as deriving accurate governing equations often requires human expert involvement. Moreover, computational efficiency remains a major obstacle, especially for high-dimensional and nonlinear PDEs, since traditional numerical solvers must manage an enormous number of equations corresponding to the system’s graph structure, often rendering these approaches prohibitively expensive~\cite{bryutkin_hamlet_2024, kumar_grade_2021, choi_gnrk_2023}.
In response to these challenges, neural differential equations (NDEs), such as Neural ODEs~\cite{chen2018neural}, have emerged as a data-driven alternative that learns the underlying dynamics directly from data, bypassing the need for explicit formulation of governing rules. This innovative approach enables the modeling of systems for which traditional equations may be intractable or unknown. Nevertheless, while NDEs excel at capturing temporal evolution, it remains challenging to model spatial dynamics, such as epidemic spread in social networks~\cite{wang2021literature} or transportation flows in urban networks~\cite{medina2022urban}, where discrete interactions complicate continuous-state representations. This limitation urgently calls for methods that can effectively integrate temporal dynamics with spatial context.


To handle the above issues, recent research has leveraged Graph Neural Networks (GNNs)~\cite{Wan_GraphODE_ICML25, scarselli2008graph, wu2019simplifying, liu2024tinygraph}, powerful tools for learning relational data, to build graph-based NDEs and model the complex interactions between variables. Early explorations integrate the graph learning capabilities of GNNs within the continuous-time framework of NDEs and propose Graph neural ODEs~\cite{poli2019graph, xhonneux2020continuous,tang2024interpretable}, which offer a versatile and powerful approach to modeling complex systems that evolve over both space and time. This integration not only enables the capture of dynamic temporal behavior but also leverages the rich spatial relationships encoded in graph structures. 
Beyond Graph Neural ODEs, the broader class of Graph Neural Differential Equations (Graph NDEs), including Graph Neural PDEs~\cite{bryutkin2024hamlet} and Graph Neural SDEs~\cite{bishnoi2024brognet}, bridges the gap between NDEs and GNNs.

\vskip 0.2em
\noindent\textbf{Contributions.}
In this work, we aim to present a comprehensive and latest review of methods that combine graph neural networks with differential equations, addressing the gap by summarizing key tasks, methodologies, and applications in this evolving field. Our contributions can be summarized as follows:
\begin{compactenum}[(a)]
    \item We offer \textit{the first comprehensive review} of Graph NDEs that model continuous spatial and temporal dynamics.
    \item We introduce a structured taxonomy of Graph NDEs in Section ~\ref{sec: tax} and conduct an in-depth review of research integrating GNNs with different classes of differential equations, including ODEs, PDEs, and SDEs, as detailed in Section ~\ref{sec: method}.
    \item We explore the diverse applications of Graph NDEs in Section ~\ref{sec: app}, highlighting their impact across various real-world scenarios.
    \item We identify emerging trends, key challenges, and promising future research directions in Section ~\ref{sec: future}, aiming to inspire further exploration in this interdisciplinary field.
\end{compactenum}

\vskip 0.2em

\noindent\textbf{Connections to existing surveys.} While previous surveys have explored Graph NDEs, they often lack comprehensiveness in methodology and categorization, limiting their ability to fully bridge GNNs and NDEs. Many focus on specific applications of neural differential equations~\cite{niu2024applications, losada2024bridging, dubeydeep}, overlooking spatial dynamics. Others examine the integration of GNNs with differential equations~\cite{han2023continuous, soleymani2024structure} but remain narrow in scope regarding DE types and categorization. In contrast, our survey compiles a broad range of recent studies, offering a detailed review of methodologies, challenges, and applications. Additionally, we present a well-structured taxonomy as well as valuable insights for future research.

\section{Background}
\subsection{Learning on Graphs}

In this paper, we define a graph as $\mathcal{G}=(\mathcal{V}, \mathcal{E})$, where $|\mathcal{V}|=N$ represents the number of nodes, and $\mathcal{E} \subseteq \mathcal{V} \times \mathcal{V}$ represents the set of edges connecting nodes. The features of all nodes is represented as $\mathbf{X}=\{\mathbf{x}_1, \mathbf{x}_2, ... \mathbf{x}_N\} \in \mathbb{R}^{N \times D}$, where $D$ denotes the feature dimension. The adjacency matrix of $\mathcal{G}$ is denoted as $\mathbf{A}$, where $\mathbf{A}_{ij}=1$ if the edge $e_{ij} \in \mathcal{E}$ and $\mathbf{A}_{ij}=0$ if $e_{ij} \notin \mathcal{E}$.
GNNs provide a flexible framework to learn graph representations. A common paradigm is message passing, where each node $v$ updates its representation $\mathbf{h}_v$ based on aggregating messages from its neighbors $\mathcal{N}(v)$. A GNN with $L$ layers can be described as:
\begin{equation}
  \mathbf{h}_v^{(l+1)} = f_\phi \Bigg(\mathbf{h}_v^{(l)},  \bigoplus_{u \in \mathcal{N}(v)} f_{\theta}\bigl(\mathbf{h}_v^{(l)}, \mathbf{h}_u^{(l)}, \mathbf{e}_{uv}\bigr)\Bigg), \forall l\in [L],
\end{equation}
where $f_\theta$ and $f_\phi$ are learnable functions parameterized by $\theta$ and $\phi$, $\mathbf{e}_{uv}$ denotes edge features (if available), and $\bigoplus$ is permutation invariant aggregation operator that aggregates neighbor information. The final representation can then be used for downstream tasks such as link prediction and graph-level classification~\cite{xie2022semisupervised}, etc.

\subsection{Neural Differential Equations}
Differential equations model dynamic systems across various domains, with their form varying based on the system. In the following, we illustrate three common types of DEs and NDEs.

\noindent\textbf{Ordinary Differential Equations (ODEs).}
ODEs describe system evolution with respect to a single independent variable, typically time \( t \). The general form is: 
$\frac{dx}{dt} = f(x(t), t)$,
where \( x(t) \) is the system state, and \( f \) dictates its rate of change.

\noindent\textbf{Partial Differential Equations (PDEs).}
PDEs involve multiple independent variables and their partial derivatives. A classical example is the diffusion equation, given by: $\frac{\partial u}{\partial t} = \alpha\nabla^2 u$, where \( u = u(x, t) \) represents the unknown quantity varying in spatial coordinate \( x \) and time \( t \), \( \alpha \) is the diffusion coefficient, quantifying the rate of spatial dispersion, and \(\nabla^2\) denotes the Laplacian operator, defined as the divergence of the gradient of the function \(u\), capturing spatial changes in systems such as fluid dynamics~\cite{belbute2020combining}.

\noindent\textbf{Stochastic Differential Equations (SDEs).}
SDEs extend ODEs by modeling the evolution of a state variable x(t) through the incorporation of randomness, often via a Wiener process \( W_t \)~\cite{bergna_uncertainty_2024}:
\begin{equation}
\label{SDE}
  dx(t)= \mu(x(t), t) dt + \sigma(x(t), t) dW_t,
\end{equation}
where \( \mu \) and \( \sigma \) are drift and diffusion terms. SDEs model systems with inherent randomness such as biological processes~\cite{liu2023group}.

\noindent\textbf{Neural Differential Equations.}
Neural Differential Equations extend classical differential equations by parameterizing the evolution function with neural networks. A prominent example is Neural Ordinary Differential Equations \cite{chen2018neural}, where a neural network models the derivative of a latent state:$\frac{d x}{d t} = f_\theta(x, t),$ where \( f_\theta \) is a neural network parameterized by \( \theta \). An ODE solver is used to compute the solution at any desired time point: $x(t) = x(t_0) + \int_{t_0}^{t} f_\theta(x(\tau), \tau) d\tau$, where $t_0$ denotes the starting time point. For back-propagation, NODEs use the adjoint sensitivity method~\cite{pontryagin2018mathematical} to solve a second ODE backward in time to compute gradients efficiently: $\frac{d a}{d t} = - a^T \frac{\partial f_\theta}{\partial x}$, where \( a = \frac{\partial L}{\partial x} \) is the adjoint state \cite{chen2018neural}. This approach enables training with constant memory cost.

\subsection{Combining GNNs with DEs}
\begin{figure}
    \centering
    \includegraphics[width=1.0\linewidth]{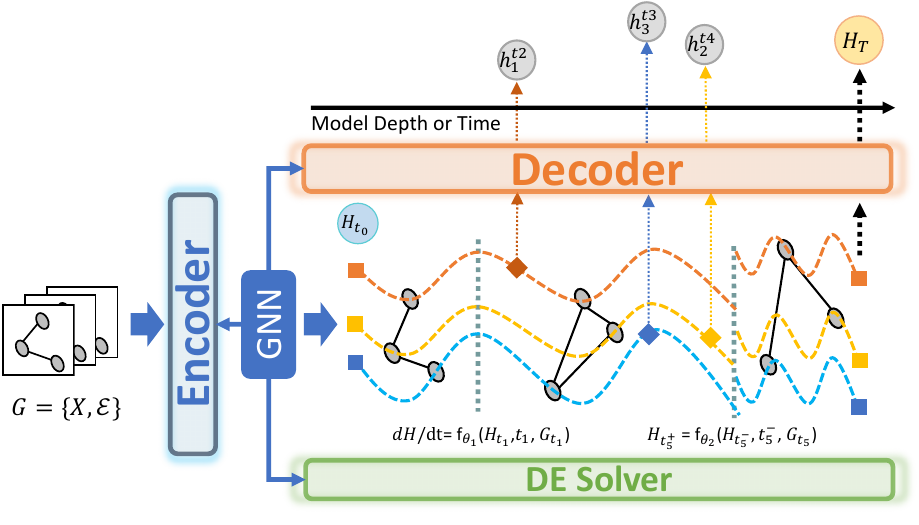}
    \vskip -1em
    \caption{GNNs can function as an \textit{encoder}, \textit{decoder}, or \textit{differential equation} in the Graph NDEs. Firstly, the encoder maps inputs to a latent initial condition, which is then propagated by the DE solver over time or model depth. Furthermore, intermediate updates can modulate state evolution through state derivatives or direct reallocation. Finally, the decoder reconstructs the latent trajectory into the target space.}
    \label{fig:framework}
    \vskip -2em
\end{figure}

By integrating the representational power of GNNs with the dynamic modeling capabilities of DEs, we introduce the concept of Graph Neural Differential Equations (Graph NDEs). A Graph NDE typically consists of two fundamental components: (i) a system of differential equations governing the temporal and spatial evolutions of states, parameterized by Neural Networks, and (ii) an initial condition that specifies the starting state of the system. While DEs establish the continuous or discrete dynamical progression of states, the role of GNNs is more flexible, as they can be incorporated at different stages of the modeling framework. As depicted in Figure~\ref{fig:framework}, Graph NDEs can be roughly categorized based on the manner in which GNNs are embedded into the system dynamics.

\subsubsection{Roles of GNNs}
\label{role of gnns}
Neural DEs generally operate in a latent space, where they model the evolution of states over time. Consequently, an encoding-decoding mechanism is typically employed: an encoder maps raw input data to a latent representation, and a decoder maps the evolved latent states into the target space. GNNs can be incorporated at various points in this pipeline, functioning as encoders and decoders and parameterizing the governing DEs.

\noindent\textbf{GNNs as Encoders.} When GNNs function as encoders, they map node feature $\mathbf{X} \in \mathbb{R}^{N \times D}$ into latent representations $\mathbf{H} \in \mathbb{R}^{N \times D'}$ while preserving relational dependencies in the graph $\mathcal{G} $~\cite{zhang_improving_2022, huang2021str, guo_evolutionary_2022, wang_causalgnn_2022}. The encoding function can be expressed as: $\mathbf{H} = GNN(\mathbf{X}, \mathcal{G}; \Theta_{enc})$, where $\Theta_{enc}$ denotes the parameters of the GNN encoder. Related works usually incorporate spatial-temporal GNNs~\cite{huang2021coupled, huang_tango_2023}, capturing both the structural and temporal information.

\noindent\textbf{GNNs as Decoders.} Given a latent representations $\mathbf{H}(t)$ evolved over time using DEs, a decoder maps it to the target space via: $\mathbf{\hat{Y}} = GNN(\mathbf{H}(t), \mathcal{G}; \Theta_{dec})$, where $\Theta_{dec}$ is the decoder parameter. Graph-based decoders enable mapping that preserves node interactions and adapts to changes to graph topology~\cite{wang_adding_2024, xunode, xiong_diffusion_2023, yan_contig_2021}.

\noindent\textbf{GNNs as Differential Equations.} Beyond serving as encoders or decoders, GNNs can be directly embedded within the differential equation to govern continuous state flows~\cite{poli2019graph, zang_neural_2020, huang2020learning, desai_variational_2021, fang_spatial-temporal_2021}. Let $x(t)$ denote the state of nodes at time $t$. The dynamics of $x(t)$ can be described by a GNN-parameterized differential equation, taking Graph Neural ODE as an example:
$\frac{dx(t)}{dt} = GNN(x(t), \mathcal{G}, \Theta)$,
where $\Theta$ is the parameter of the GNN that models continuous aggregation across the graph. Similarly, higher-order formulations extend this principle as:
$\frac{d^kx(t)}{dt^k} = GNN(x(t), \mathcal{G}, \Theta)$,
where $k$ denotes the order of the DE. Such formulation enables enhanced expressiveness and adaptability in evolving graph structures~\cite{iakovlev_learning_2021, kumar_grade_2021, rusch_graph-coupled_2022, luo_hope_2023, niwa_coordinet_2023, wen_second-order_2024, liu_segno_2024, eliasof_temporal_2024, huang_link_2024}.

\subsubsection{Initial Condition Construction}
The definition of the initial condition $x(t_0)$ significantly influences the trajectory of the learned dynamics. Besides latent encodings, initial states can be derived from raw inputs or learned embeddings.

\noindent\textbf{Encoding-Based Initialization.} In this case, raw node features $\mathbf{X}$ are first mapped into a latent space before being used as the latent initial condition, which can be deterministic: $\mathbf{H}(t_0) = f(\mathbf{X}, \mathcal{G}; \Theta_{enc})$ or sampled from a given distribution~\cite{huang2020learning, huang2021coupled, gao_egpde-net_2023, huang_tango_2023} (e.g. Gaussian): $\mathbf{H}(t_0) \thicksim \mathcal{N}(\hat{\mathbf{\mu}}, \hat{\mathbf{\sigma}})$, where $\hat{\mathbf{\mu}}$ and $\hat{\mathbf{\sigma}}$ are inferred by a function $f(\mathbf{X}, \mathcal{G})$, introducing stochasticity into the initial states to model uncertainty.

\noindent\textbf{Pre-defined Initialization.} In physical systems, the initial condition is often dictated by domain constraints, leading to $\mathbf{H}(t_0) = \mathbf{X}$, where no additional encoding is applied. This approach is common in dynamic simulations~\cite{kosma_neural_nodate, uwamichi_integrating_2024, yuan2024egode}, where initial states are predefined or randomly generated.

\noindent\textbf{Learning-Based Initialization.} Instead of explicitly defining initial states and encoding raw features, it is also viable to learn the initial condition during model training. The model learns an optimal embedding $\mathbf{H}(t_0)$ that best facilitates downstream tasks: $ \mathbf{H}(t_0) = \text{argmin}_{\mathbf{H}(t_0)} \mathcal{L}(\mathcal{F}(\mathbf{H}(t_0), \mathcal{G}))$, where $\mathcal{L}$ represents a task-specific objective function and $\mathcal{F}$ denotes the differential equation dynamics. This strategy is particularly effective in recommendation systems~\cite{choi2021lt, yang2024siamese, han2021learning}, where graph structure alone is available, and node representations must be inferred from relational interactions.

\section{Taxonomies}
\label{sec: tax}
\begin{figure*}
    \centering
    \includegraphics[width=0.9\linewidth]{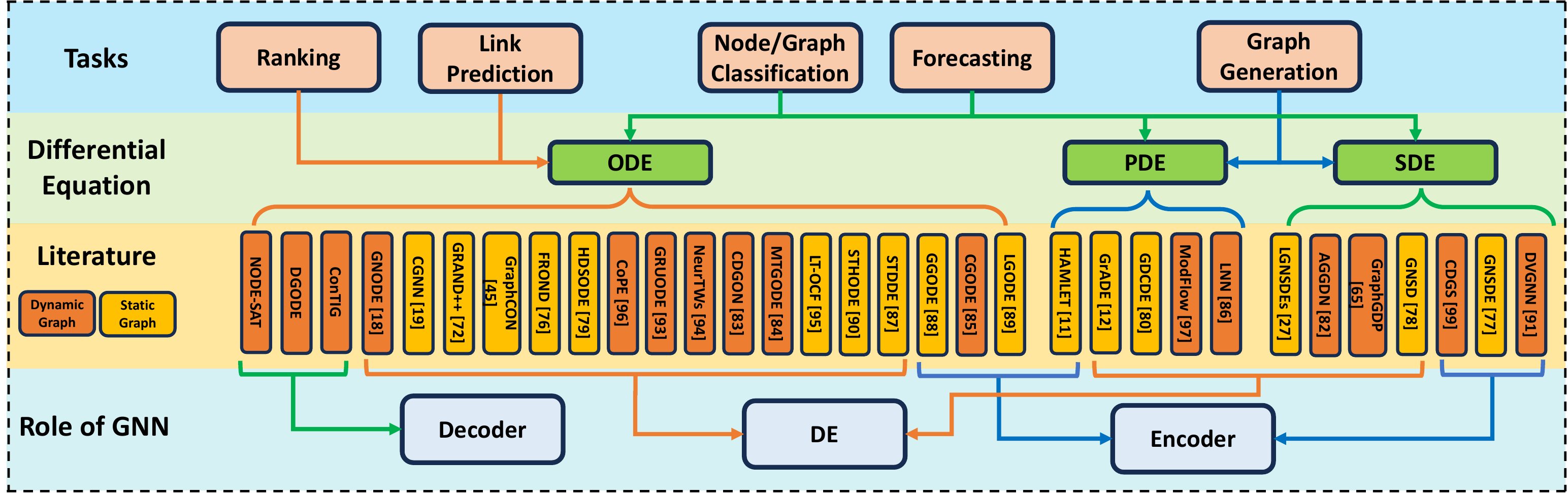}
    \vskip -1em
    \caption{Summary of Graph Neural Differential Equation methods.}
    \label{fig: summary}
    \vskip -1em
\end{figure*}

We classify Graph NDEs based on tasks, datasets, graph construction techniques, and methodological distinctions. A complete categorization is in Appendix~\ref{appendix: A}, with a partial version in Figure~\ref{fig: summary}.

\subsection{Tasks}
Among all the research investigated in this paper, Graph NDEs are applied in five primary tasks: \textit{Node/Graph Classification}, \textit{Link Prediction}, \textit{Ranking}, \textit{Forecasting}, and \textit{Graph Generation}.


\noindent\textbf{Node/Graph Classification.} Given $\mathcal{G} = (\mathcal{V}, \mathcal{E})$, the goal is to learn functions $f:\mathcal{V} \to \mathcal{Y}$ (node classification) or $g:\mathcal{G} \to \mathcal{Y}$ (graph classification). Graph NDEs model message passing as a continuous process rather than a one-step discrete propagation. For citation network prediction~\cite{wang_acmp_2023}, articles (nodes) and citations (edges) form citation graphs, where Graph NDEs improve representation learning by capturing continuous citing patterns.



\noindent\textbf{Link Prediction.}
Given $\mathcal{G} = (\mathcal{V}, \mathcal{E})$, link prediction learns $h: \mathcal{V} \times \mathcal{V} \to [0,1]$ to estimate edge existence probability. Graph NDEs enhance prediction by modeling continuous node embedding dynamics. Such a task is commonly seen in recommendation systems and knowledge graphs~\cite{huang_link_2024, han_learning_2021}.


\noindent\textbf{Ranking.}
Ranking assigns scores to nodes, optimizing $s: \mathcal{V} \to \mathbb{R}$ for ordered retrieval. Graph NDEs leverage continuous diffusion to model information propagation. For example, in recommendation systems~\cite{guo_evolutionary_2022}, users and items form bipartite graphs; Graph NDEs capture evolving preferences, refining interaction prediction.


\noindent\textbf{Forecasting.}
For dynamic graphs $\mathcal{G}(t) = (\mathcal{V}, \mathcal{E}(t))$ with temporal node features $\mathbf{X}(t)$, forecasting estimates future states via $f: \mathcal{V} \times \mathbb{R}^{s} \to \mathbb{R}^{z}$, where $s$ denotes the history input size and $z$ refers to the horizon of prediction. Graph NDEs, incorporating time-continuous dynamics, excel in capturing gradual state transitions. For example, in traffic flow forecasting~\cite{fang_spatial-temporal_2021, wu_continuously_2024, ma_spatio-temporal_2024, zhong_attention-based_2023}, road networks use Graph NDEs for real-time flow updates and long-term forecasting.


\noindent\textbf{Graph Generation.}
Given a training set ${\mathcal{G}_i}$, graph generation models $p(\mathcal{G})$ to sample structurally meaningful graphs. Graph NDEs enable continuous latent space exploration, improving structural coherence and diversity in generated samples~\cite{noauthor_conditional_nodate, huang_graphgdp_2022}.

\subsection{Graph Construction}
The construction of graphs plays a crucial role in shaping both the design and performance of Graph NDEs. This process can be analyzed along two key dimensions: \textit{spatial} and \textit{temporal}.

\vskip -4em
\subsubsection{Spatial Level.} In this paper, we depict \textit{spatial-level relations} as a general notion of proximity or relationships between nodes. Depending on how nodes and edges are formulated, graph construction typically follows one of two primary approaches:

\noindent\textbf{Point-Based Graphs.} Point-based graphs can be irregular, where nodes correspond to individual data points, and edges are established based on a function $f_e$: $\mathcal{V} \times \mathcal{V} \to \{0,1\}$ that determines connectivity based on proximity from observations. The notion of proximity can be defined in various ways depending on the nature of the data, often reflecting spatial~\cite{klemmer2023positional}, or semantic relationships~\cite{ma2023language} among nodes. To quantify the strength of connectivity, edge weights can also be applied.

\noindent\textbf{Grid-Based Graphs.} A grid-based graph is a spatially regular, where $\mathcal{V} \subset \mathbb{Z}^n$ represents nodes positioned at integer lattice points in an $n$-dimensional space, and edges $\mathcal{E}$ connect nodes based on a predefined neighborhood structure. This structured representation is widely used in drone swarming~\cite{jiahao_learning_2021}, and physical modeling~\cite{iakovlev_learning_2021, bryutkin_hamlet_2024, kumar_grade_2021, choi_gnrk_2023}, where data is arranged in a spatially regular manner.


\subsubsection{Temporal Level.}  
Temporal graphs evolve over time in terms of node features or graph topology. Therefore, each time point yields a distinct graph \(\mathcal{G}(t) = (\mathcal{V}(t), \mathcal{E}(t))\). Formally, temporal graphs can be categorized into:

\noindent\textbf{Static Graph.} The temporal evolution of a static graph \(\mathcal{G} = (\mathcal{V}, \mathcal{E})\) is solely captured through time-dependent node attributes \(\mathbf{X}(t)\). That is, each node \(v \in \mathcal{V}\) has a feature vector \(\mathbf{x}_v(t)\) evolving over time, while the edge set remains unchanged, i.e., \(\mathcal{E}(t) = \mathcal{E}, \forall t\). Applications span citation~\cite{poli2019graph} and traffic~\cite{zhong_attention-based_2023} networks.

\noindent\textbf{Dynamic Graph.} A dynamic graph is characterized by time-evolving edges and edge weights, \(\mathcal{E}(t)\), meaning both connectivity and interaction strengths change over time. The structure of such graphs is determined via: i) explicit modification of adjacency relations, where \(\mathcal{E}(t) = \{ e_{uv}(t) \}\) updates based on new inputs; ii) adaptive learning of edges and weights during training, where adjacency matrices are replaced by a learned attention matrix \( A_t = (a_{ij}) \in \mathbb{R}^{N \times N} \), with \( a_{ij} = f(v_i, v_j, t) \) capturing dynamic influence. The temporal evolution of node states follows Graph NDEs: $\frac{dx(t)}{dt} = f_{\theta}(x(t), \mathcal{E}(t))$, which model node state evolution under varying graph structure. Applications span social interactions~\cite{zhang_improving_2022}, graph generation~\cite{noauthor_conditional_nodate, huang_graphgdp_2022}, etc.

\subsection{Modeling Spatial \& Temporal Dynamics}
\label{sec: graph NDEs}
Graph NDEs surpass discrete models by providing a continuous flow of latent states across spatial and temporal dimensions, and the key lies in the modeling of \textit{spatial} and \textit{temporal dynamics}.

\noindent\textbf{Temporal Dynamics Modeling.}
\label{sec: temporal_dynamics}
In classical NDEs, the continuous evolution of variable states is defined with respect to actual time, \( t \in \mathbb{R}^{+} \), ensuring alignment with the target trajectory~\cite{chen2018neural}. Specifically, in Graph NDEs, the state of each node, \( x(t) \), evolves according to a time-dependent differential equation $\frac{\mathrm{d} x(t)}{\mathrm{d} t} \;=\; f_\theta\!\bigl(x(t), t\bigr)$, where the graph structure is either encoded in the initial condition \( x(t_0) \)~\cite{huang2021str} or integrated into the function \( f_\theta \). Additionally, temporal dynamics are influenced by external controls or new inputs, which can update node states and introduce new graph structures, thereby altering the flow of states over time. Spatial-temporal models extend NDEs by explicitly incorporating the spatial dimension, capturing both spatial and temporal evolution on dynamic graphs. 

\noindent\textbf{Spatial Dynamics Modeling.}
\label{sec:spatial_dynamics}
The dynamic spatial evolution can also be applied to static graphs, where the model \textit{depth}~\cite{xhonneux2020continuous} corresponds to a continuous notion of time. Unlike the conventional approach of stacking discrete GNN layers, this continuous perspective naturally connects to diffusion equations~\cite{chamberlain_grand_2021}. A general graph diffusion equation is given by:
\begin{equation}
    \frac{\partial x(t)}{\partial t} \;=\; \mathrm{div} \Bigl(\,G\bigl(x(t), t\bigr) \cdot \nabla x(t)\Bigr),
\end{equation}
where \(\nabla x(t)\) is the divergence of \(x\), \(G\bigl(x(t), t\bigr)\) is the diffusion coefficient that may depend on both the current state \(x(t)\) as well as current time or depth \(t\), and \(\mathrm{div} (\cdot)\) denotes the graph divergence operator. By parameterizing this diffusion process, we arrive at the formulation of NDEs on graphs.

Recent work provides further insight into this connection. For instance, Chamberlain et al.~\cite{chamberlain_beltrami_2021} show that GNNs can be viewed as the discrete form of Beltrami flow, while Choi et al.~\cite{choi_gread_2023} develop a reaction-diffusion-based GNN architecture. Taken together, these studies reinforce the link between Graph NDEs and graph diffusion processes. Consequently, Graph NDEs provide both smoother feature propagation across the graph and a principled physical analogy grounded in a well-established diffusion process.

\section{Methodology}
\label{sec: method}
In this section, we detail the methodology underpinning Graph Neural Differential Equations (Graph NDEs). Our discussion is organized around two primary perspectives: \textit{Temporal Dynamics Modeling} and \textit{Spatial Dynamics Modeling}. For each perspective, we elaborate the unique challenges involved.

\subsection{Temporal Dynamics Modeling}
For spatial-temporal models, incorporating the time dimension presents several challenges. This section outlines key temporal modeling challenges and corresponding solutions in Graph NDEs, including \textit{dynamic updates}, \textit{irregular time intervals}, \textit{modeling temporal delay}, \textit{modeling hybrid system dynamics}, and the \textit{efficiency}.


\subsubsection{Dynamic Temporal Updates}
\label{temporal updates}

For a naive DE, whether parameterized by an NN or not, a fixed trajectory is predicted once the initial condition is given. This is because the DE defines the evolution or flow of node states, namely the vector field~\cite{chen2018neural}. However, in a spatial-temporal graph, both node features and the graph structure can evolve over time independently of the vector field defined by the DE. These updates significantly impact the target trajectory. From the perspective of the vector field, the updated graphs can be interpreted in two ways: 1. Reallocation of states, and 2. Conditioned states flow. For the reallocation of states, the vector field does not change while the new inputs introduce a jump of states in the vector field: $x\bigl(t_k^+\bigr) \;=\; \Phi\Bigl(x\bigl(t_k^-\bigr), \Delta(t_k)\Bigr)$,
where \(t_k\) denotes the moment of a discrete update, \(t_k^-\) and \(t_k^+\) are the times just before and just after the jump, respectively, and \(\Delta(t_k)\) captures the new information or structural change at \(t_k\). The update function \(\Phi\) then adjusts the node states accordingly. This jump effectively resets the trajectory of the node states, altering the dynamics governed by the DE. For example, Poli et al.~\cite{poli2019graph} introduced an autoregressive graph differential equation that applies "jumps" to adapt to the dynamic graph structure., and Zhang et al.~\cite{zhang_cope_2021} applied Graph NDE in the recommendation system, where jumps of node features are introduced as the interaction between the user and item changes.

In the case of a conditioned state flow, newly arrived inputs at a given time point serve as the input of Graph NDEs, which gives a formulation of $ \frac{dX}{dt} = F( \mathbf{X}(t), \mathcal{G}(t))$, where $\mathbf{X}(t)$ is the current node states and $\mathcal{G}(t)$ is the dynamic graph including node features and the graph structure. The change of $\mathcal{G}(t)$ can be viewed as either a change of the vector field or as an adjustment to the output $ \frac{dX}{dt}$. Instead of solely relying on the initial condition, the vector field becomes conditioned on the current input, allowing the system to adapt its trajectory in response to external influences~\cite{xing_aggdn_2024, jin_multivariate_2023}.

\subsubsection{Irregular Time Interval.}
Real-world dynamic systems exhibit irregularly sampled time series, where observations occur at non-uniform intervals, making traditional discrete-time models ineffective. DE-based models naturally address this by modeling the continuous evolution of node states so that the states at arbitrary time points can be inferred.
Recent advancements, such as LG-ODE~\cite{Huang2020LGODE}, CG-ODE~\cite{Huang2021CGODE}, and GG-ODE~\cite{Huang2023GGODE}, extend such ability to graph-structured data. LG-ODE~\cite{Huang2020LGODE} models continuous node dynamics using latent ODEs, enabling interpolation across uneven time steps. CG-ODE~\cite{Huang2021CGODE} further generalizes this by incorporating evolving graph structures, where both node states and edge interactions are learned through coupled differential equations. GG-ODE~\cite{Huang2023GGODE} extends these ideas across multiple environments by introducing environment-specific latent factors, enabling the transfer of learned dynamics across different systems. These approaches provide a flexible framework for modeling real-world graph dynamics under irregular sampling, outperforming discrete-time methods in handling asynchronous, partially observed data.

\subsubsection{Temporal Delay Modeling.}
Traditional GNNs assume immediate information propagation, which fails to capture the inherent temporal delays present in real-world systems. In applications such as traffic forecasting, changes in one location take time to influence others, making delay-aware modeling essential. Long et al.~\cite{long2024unveiling} introduces the Spatial-Temporal Delay Differential Equation, which explicitly incorporates time delays into spatial-temporal modeling. The core idea is to model node interactions as: $\frac{d h_i(t)}{dt} = f(h_i(t), h_j(t - \tau_{ij}), \theta)$, where $\tau_{ij}$ represents the time delay in information propagation between nodes. Instead of assuming fixed delays, they propose two approaches: (1) a precomputed delay estimator using \textit{max-cross correlation}, and (2) a time-delay estimator that dynamically adjusts $\tau_{ij}$ based on traffic conditions.





\subsubsection{Temporal Dynamics for Hybrid Systems.}
While Graph NDEs parameterize the DEs with NNs or GNNs, making the model inherently data-driven, incorporating domain-specific biases can be beneficial for regularizing the model outputs. Such modeling biases, often derived from well-established physical laws, constraints, or expert knowledge, are fundamental in traditional knowledge-driven models. To enhance the effectiveness and interpretability of Graph NDEs, hybrid models~\cite{liu2024review} integrate domain-knowledge into the modeling of Graph NDEs by either \textit{constraining the form of Graph NDEs} or \textit{predicting the abstract quantities or parameters of the physical system}. Li et al.~\cite{li_physics-informed_2024} extend the use of the spatial-temporal decay model from one-dimensional dynamics to the high-dimensional latent space. Similarly, Han et al.~\cite{han2023devil} make use of the structure of the Susceptied-Infected-Recovered model and switch the modeling to the latent space. On the other hand, Sanchez-Gonzalez et al.~\cite{sanchez2019hamiltonian} model the Hamiltonian mechanism by predicting the momentum and velocity in the physical function using a GNN. Similar practices have been made for Lagrangian mechanism~\cite{cranmer2020lagrangian}.

\subsubsection{Efficient Temporal Simulation}  
Although classic PDE models effectively describe various real-world phenomena with numerical solvers such as the Finite Element Method~\cite{cecil1977finite}, time complexity remains a significant challenge, particularly for solving complex dynamics and real-time processing tasks~\cite{jiang_neural_2023}. The primary sources of this challenge are twofold. First, the high dimensionality and non-linearity of many problems lead to more intricate PDE systems, both in terms of the number of equations and their structural complexity. Second, the demand for larger or more fine-grained grid-based graphs further increases the number of nodes to be modeled~\cite{han2013numerical}, which can be computationally expensive using classic PDE solvers. Beyond the computational inefficiency of handling large-scale and complex PDEs, additional challenges arise when dealing with problems that lack a fixed PDE formulation. To address these issues, Graph Neural PDEs~\cite{iakovlev_learning_2021, bryutkin_hamlet_2024, kumar_grade_2021, choi_gnrk_2023} employ a data-driven approach to learn and integrate the governing rules of both spatial and temporal dynamics across all nodes using a single model. Unlike traditional methods, where computational complexity scales with problem intricacy and the graph size, Graph Neural PDEs maintain a fixed model for all cases and does not rely on classic PDE solvers. As a result, the time complexity does not increase with the problem complexity, and the governing dynamics are encapsulated within the learned parameters, enabling efficient and scalable solutions.


\subsection{Spatial Dynamics Modeling}
Unlike spatial-temporal dynamics which model latent state evolution over time, spatial dynamics can also be modeled in terms of model depth. While traditional GNNs effectively capture static relationships through local aggregation, their discrete nature constrains their ability to represent continuous feature evolution. Here, we show how embedding GNNs in differential equations enables dynamic graph modeling and overcomes classic challenges such as \textit{over-smoothing}, \textit{measuring uncertainty}, \textit{adversarial robustness}, \textit{graph heterophily}, and \textit{modeling high-order relations}.

\subsubsection{Over-Smoothing on Graphs.}

GNNs with deeper architecture experience severe performance degradation due to vanishing gradient and over-smoothing, where node representations become indistinguishable and converge to the same value as more layers are added ~\cite{rusch2023survey}. To mitigate the effect of over-smoothing, previous studies imitate residual networks~\cite{he2016deep} and develop skip connections~\cite{xu2021optimization} between layers, updating the node features using $\textbf{H}(t+1)=\textbf{H}(t)+\text{GNN}_\theta(\textbf{H}(t), \mathcal{G})$, where t denotes the layer index. On the other hand, Graph NODE~\cite{poli2019graph} takes a step further by making this process differential: $\frac{d\textbf{H}(t)}{dt} = \text{GNN}(\textbf{H}(t), \mathcal{G})$. Then, a numerical solver~\cite{finlay2020train} is applied to acquire the trajectory. Additionally, CGNN ~\cite{xhonneux2020continuous} introduces the initial latent embeddings,  $E = \textbf{H}(t_0)$, into the ODE formulation: $\frac{d\textbf{H}(t)}{dt} = \text{GNN}(\textbf{H}(t), \mathcal{G}) + \textbf{H}(t_0)$ . This results in a function with a restart distribution, which helps the model retain the initial representations and effectively mitigates over-smoothing. Moreover,
GRAND++~\cite{thorpe_grand_2022} interprets over-smoothing through the lens of diffusion, where deeper networks excessively diffuse node features, ultimately leading to uniform feature representations across all nodes. Similar to CGNN, Grand++ mitigates this issue by introducing a source term (restart) in the differential equation (DE) to preserve initial representations. Likewise, Graph-Coupled Oscillator Networks~\cite{rusch_graph-coupled_2022} establish a connection between over-smoothing and zero-Dirichlet energy steady states, proposing a second-order ODE to counteract the over-smoothing. Furthermore, Maskey et al. ~\cite{maskey_fractional_nodate} extend the problem to directed graphs and tackle over-smoothing with fractional graph Laplacians.

\subsubsection{Uncertainty within Graph Dynamics.}
Real-world data often contains noise and unobserved external factors that influence the dynamics of graph propagation. For one thing, graphs are often constructed from real-world data where the connections (edges) between nodes can be incomplete, noisy, or even spurious~\cite{hao2021ks}. For another thing, even though a clear graph structure is given, the way information, influence, or any form of signal diffuses through these networks can be highly variable and subject to external factors (e.g., weather, human behavior)~\cite{wu2024graph}. Since GNNs and Graph NODEs make predictions conditioned on neighbors, both kinds of uncertainty impact their performance. To provide a direct measurement of uncertainty and improve the robustness of these models, Graph Neural SDEs introduce a stochastic diffusion term $\sigma(x(t), t) dW_t$, as illustrated in Equation~\ref{SDE}, enhancing the performance on node classification tasks in both the In Distribution and Out of Distribution cases~\cite{bergna_graph_2023, bergna_uncertainty_2024, lin2024graph}. Furthermore, Liang et al.~\cite{liang2024dynamic} combines graph variational encoding with SDE, generating dynamic graphs for spatial-temporal forecasting. Xing et al.~\cite{xing_aggdn_2024} stacked the SDE module upon ODE, which works as a control signal to modulate the SDE propagation. Huang et al. ~\cite{huang_graphgdp_2022} applied SDE in the graph generation task by applying the reverse-time SDE to generate the target permutation-invariant graphs from random graphs.

\subsubsection{Graph Adversarial Robustness}
GNNs are vulnerable to adversarial perturbations due to inter-node information exchange. Adversaries can perform modification attacks by adding or removing edges or injection attacks by introducing malicious nodes.
Song et al.~\cite{song2022robustness} treat graphs as discretized Riemannian manifolds and analyze the stability of the heat kernel under metric perturbations. Their results show that for small perturbations \(\varepsilon = o(1)\), the change in node features remains bounded \(\bigl\|\varphi(u, t) - \tilde{\varphi}(u, t)\bigr\| = O(\varepsilon)\), where $\varphi(u, t)$ is the node attribute of node $u$ at time $t$, indicating that PDE-based GNNs can better withstand adversarial topology attacks. Building on this analysis, they propose a novel class of graph neural PDEs with stronger defenses against such adversarial modifications.
While Song et al. ~\cite{song2022robustness} demonstrate Lyapunov stability, it does not necessarily guarantee adversarial robustness. Zhao et al. ~\cite{zhao2024adversarial} analyze various stability concepts for graph neural flows, leading to the Hamiltonian Graph diffusion class, which improves robustness by maintaining constant total Hamiltonian energy over time, ensuring bounded BIBO stability. Recently, Kang et al. ~\cite{kang2024coupling} made an extension to graph neural fractional-order differential equations, showing more robust than existing Graph neural ODEs.

\subsubsection{Graph Heterophily}
GNNs have been widely used for various graph-based learning tasks, yet they often assume connected nodes have similar attributes (homophily), which is not hold in heterophilic graphs, leading to suboptimal performance. To address this, recent works have explored Neural ODEs combined with graph dynamic modeling to enhance node representation learning in heterophilic settings.
Recent studies have introduced diffusion-based models to handle heterophilic graphs effectively. Zhao et al. ~\cite{zhao2023graph} propose a Graph Neural Convection-Diffusion framework, leveraging the convection-diffusion equation (CDE) to incorporate both homophilic and heterophilic information. The convection-diffusion equation is formulated as: $\frac{\partial x}{\partial t} = \text{div}(D\nabla x) - \text{div}(v x)$,
where the first term represents diffusion, and the second term accounts for convection with velocity field \( v \) controlling the propagation direction. In the discrete graph setting, this extends to: $\frac{\partial x(t)}{\partial t} = \text{div}(D(x(t), t) \odot \nabla x(t)) + \text{div}(v(t) \circ x(t))$,
where \( v(t) \) now adapts to node dissimilarity, enhancing classification performance on heterophilic graphs. Similarly, Zhang and Li ~\cite{zhang2024unleashing} introduce a dual-channel Continuous Graph Neural Network with latent states applied using low-pass ($\textbf{H}_L$) and high-pass ($\textbf{H}_H$) filtering :
\begin{equation}
    \frac{\partial \textbf{H}_L}{\partial t} = (\hat{{\bf A}}_{\text{sym}} - {\bf I}) \textbf{H}_L + \textbf{H}(t_0), \quad
    \frac{\partial \textbf{H}_H}{\partial t} = (-\hat{{\bf A}}_{\text{sym}}) \textbf{H}_H + \textbf{H}(t_0).
\end{equation}
where $\hat{{\bf A}}_{\text{sym}}$ is the symmetrically normalized adjacency matrix, and the features are mixed from both channels in the end. 
\subsubsection{Graph Dynamics with High-Order Relations.}
Many real-world problems involve interactions that go beyond pairwise relationships. Hypergraph learning~\cite{feng2019hypergraph} addresses this by allowing each hyperedge to connect multiple nodes.  Nevertheless, incorporating hypergraph learning into the framework of Graph NDEs is non-trivial as the dynamics of pair-wise graphs and hypergraphs are different. To bridge the gap, Yao et al.~\cite{yao_spatio-temporal_2023} propose to model spatial and temporal evolutions separately, building a spatial hypergraph $G_{sp}$ and a temporal hypergraph $G_{te}$. Then, hypergraph convolution is integrated into the ODE, which yields the spatial and temporal evolutions on the two graphs. In the end, an MLP layer is applied to combine the final embeddings from the spatial and temporal levels.
Besides separately encoding the spatial and temporal evolutions, Yan et al.~\cite{yan_hypergraph_2024} propose to encode the node embeddings $\mathbf{H}_v$ and hyperedge embeddings $\mathbf{H}_e$ separately, which gives an ODE in the form of:
$ \begin{bmatrix}\dot{\mathbf{H}}_v \\ \dot{\mathbf{H}}_e \end{bmatrix} = \begin{bmatrix} g_v\bigl(\mathbf{H}_v(t) \bigr) \\ g_e\bigl(\mathbf{H}_e(t)\bigr) \end{bmatrix} \;+\; A \begin{bmatrix} \mathbf{H}_v(t) \\ \mathbf{H}_e(t) \end{bmatrix}$,
where $g_v$ and $g_e$ are the control functions and $A$ denotes the diffusion velocity effect between the vertex representation and the hyperedge representation in the dynamic system by the correlation of the hypergraph. 


\section{Applications}
\label{sec: app}
Graph NDEs have been applied across various domains due to their capability to model continuous spatial and temporal dynamics.
In this section, we discuss on some popular applications, including \textit{Physics Systems Simulation}, \textit{Traffic Flow Forecasting},  \textit{Recommendation Systems},  \textit{Epidemic Modeling}, and  \textit{Graph Generation}.

\subsection{Physics Systems Simulation}

Graph Neural ODEs have proven effective for modeling continuous-time dynamics in physics-based simulations by parameterizing system evolution through differential equations, allowing flexible and efficient trajectory prediction. These models are particularly useful for simulating multi-body interactions~\cite{Huang2023GGODE}, particle dynamics~\cite{shi2024towards}, spring system~\cite{huang2020learning}, charged particle system~\cite{huang2020learning}, chaotic pendulum system~\cite{huang2024physics} and fluid mechanics~\cite{wu2025pure}, where relational structures naturally fit graph representations.
For example, the Hamiltonian Graph Network \cite{Sanchez-Gonzalez2019HGN} incorporates Hamiltonian mechanics to enforce energy conservation in learned physics models. GG-ODE \cite{Huang2023GGODE} introduces environment-specific latent factors to adapt physics models across different conditions. EGODE \cite{yuan2024egode} extends Graph Neural ODEs to hybrid systems, handling sudden state changes, e.g.,rigid-body collisions. Furthermore, GNSTODE \cite{Shi2023GNSTODE} improves spatial-temporal modeling in physics systems by learning latent force interactions and refining long-range dependencies. These models demonstrate how Graph ODEs enhance the accuracy, efficiency, and generalization of physics-based simulations, outperforming numerical solvers in long-term stability and adaptability.

\subsection{Traffic Flow Forecasting}

Traffic flow forecasting is a crucial task in intelligent transportation systems, requiring models that can capture complex spatial-temporal dependencies. Traditional models, such as ARIMA and LSTM-based approaches, struggle with irregular traffic patterns and evolving road network dynamics. Graph NDEs provide a continuous-time framework that integrates spatial-temporal dynamics, improving long-range forecasting and adapting to variable time intervals.
For example, STGODE \cite{fang2021stgode} models traffic flow as a continuous dynamical system, integrating GNNs with an ODE solver to handle long-term dependencies. GODE-RNN \cite{su2022gode_rnn} combines Graph NDEs with RNNs, capturing both fine-grained temporal changes and spatial interactions. ASTGODE \cite{zhong2023astgode} introduces an attention mechanism within Graph ODEs to enhance interpretability and adaptive forecasting. Additionally, GRAM-ODE \cite{liu2023gramode} employs multiple Graph ODE modules to learn hierarchical traffic patterns, while AGODE \cite{bai2024agode} dynamically updates the graph structure to reflect changing traffic conditions. These methods outperform discrete-time GNN models in accuracy, demonstrating the potential of Graph NDEs to handle irregular and evolving traffic data efficiently.

\subsection{Recommendation Systems}
Recommendation systems~\cite{da2020recommendation} naturally form a bipartite graph structure, capturing relationships between users and items. Graph NDEs effectively model the continuous evolution of user preferences, surpassing traditional collaborative filtering methods~\cite{su2009survey} by accounting for dynamic interactions. 
For example, Qin et al.~\cite{qin_learning_2024} propose an autoregressive propagation framework with an edge-evolving mechanism and a temporal aggregation module to predict user-item interactions, which is similar to CoPE~\cite{zhang_cope_2021} and ConTIG~\cite{yan_contig_2021}. To further enhance the learned representations of Graph NDEs, Yang et al.~\cite{yang_siamese_2024} integrate contrastive learning into their optimization process. Additionally, to improve adaptability to dynamic graphs, Guo et al.~\cite{guo_evolutionary_2022} propose t-Alignment, which synchronizes the updating time steps of temporal session graphs within a batch.

\subsection{Epidemic Modeling}
Modeling infectious disease spread~\cite{duan2015mathematical,liu2024review,liu2024epilearn} is crucial for public health policy design. In an epidemic graph, nodes represent individuals or communities, while edges denote their interactions. Graph NDEs extend beyond the traditional mechanistic models, e.g., SIR and its variants~\cite{rodriguez2022data}, introducing \textit{hybrid models}~\cite{liu2024review} that integrate both GNNs and mechanistic models. These approaches enhance the ability to capture complex infection dynamics. 
For example, STAN~\cite{gao2021stan} preserves the ODE function of the SIR model and uses GNNs to predict the parameters of the SIR model on a static graph. As an extension, MepoGNN~\cite{cao2022mepognn} adopts a graph learning module, which introduces learning on dynamic graphs. Besides learning the parameters of mechanistic models, Wan et al.~\cite{wan_epidemiology-aware_2024} integrate the ODE function of the SIR model with GNNs, modeling the variables in the high-dimensional latent space.

\subsection{Graph Generation}  
Graph generation is essential for applications like drug discovery and program synthesis~\cite{zhu2022survey}, but modeling graph distributions is challenging due to their discrete, permutation-invariant nature. Traditional models like variational autoencoders~\cite{doersch2016tutorial} struggle with this invariance. Score-based generative models address this by using graph SDEs to simulate graph trajectories, where diffusion corrupts graphs into a prior distribution (e.g., normal distribution), and the trajectory captures both diffusion and denoising. These models rely on log-density gradient vector fields, imposing fewer constraints than likelihood-based approaches while ensuring permutation invariance. The edge-wise dense prediction GNN~\cite{niu2020permutation} estimates scores for graph distributions while maintaining permutation invariance but is limited to adjacency matrices. To overcome this, Jo et al.~\cite{jo2022score} proposed Graph Diffusion via SDEs, which models both node features and adjacency matrices with separate drift and diffusion terms, capturing node-edge dependencies. CDGS~\cite{huang2023conditional} further improves this with hybrid message-passing blocks and fast ODE solvers, enabling rapid, high-quality molecule generation.

\section{Future Work}
\label{sec: future}

While significant advancements in Graph NDEs, many challenges remain largely unexplored. In this section, we discuss these issues and suggest directions for future research.


\subsection{Discovering Graph Differential Equations}
Equation discovery is an essential task across scientific disciplines, facilitating the extraction of explicit mathematical relationships directly from observed data~\cite{angelis2023artificial}. At the heart of this process lies \textit{symbolic regression}. Recently, deep learning-based methods, such as set-to-sequence transformers~\cite{biggio2021neural} and large language models~\cite{shojaee2024llm}, have emerged as viable alternatives to traditional symbolic regression approaches. However, due to inherent architectural constraints, they focus on discrete representations of data. In contrast, Graph NDEs explicitly capture continuous dynamical systems by learning vector field representations, yet they typically encode equations in implicit forms. This fundamental difference indicates that integrating discrete-focused transformer-based methods with continuous-based Graph NDEs presents a compelling pathway toward advancing the field of differential equation discovery.

\subsection{Handling Graph Sparsity and Sporadicity}
Data sparsity in dynamic systems remains a critical challenge, often manifesting as limited labeled nodes and missing or incomplete observations over time and space. 
Additionally, sporadic patterns, characterized by both sparsity and irregular, unpredictable distribution, further challenge learning and inference~\cite{yan2024enhanced}. Recently, Luo et al.~\cite{luo2023graph} combined the strengths of neural processes and neural ODEs to model evolving graphs with missing edges and to capture physical dynamics from highly sparse spatial-temporal data. Nevertheless, open challenges persist in ensuring the robustness and efficiency of Graph NDEs under extreme sparsity conditions. To further address the challenge, possible solutions may involve zero-shot annotator to label a small portion of nodes~\cite{chen2023label}, or graph condensation~\cite{hashemi2024comprehensive} that yields a condensed graph from sparse graph.


\subsection{Scalability on Large Dynamic Graphs}
Scalability remains a significant challenge for Graph NDEs, particularly when applied to large-scale dynamic graphs. These models require the solving of continuous-time differential equations for potentially millions of nodes and edges, leading to high computational overhead and memory demands~\cite{finzi2023stable}. The iterative nature of DE solvers exacerbates the issue since repeated evaluations of neural network functions over numerous time steps can be prohibitively time-consuming. 
To tackle this challenge, several approaches can be explored: developing efficient numerical solvers tailored for neural differential equations~\cite{han2018solving}, leveraging parallelization and GPU acceleration~\cite{samat2021gpu}, and employing sparse representations and approximation methods~\cite{jagtap2020conservative}.

\subsection{Modeling Continuous Structural Evolution}
Real-world graphs, such as social networks, frequently undergo dynamic changes, with nodes and edges continuously evolving over time. While several studies have introduced flexible approaches to incorporate dynamic inputs during inference, either by adjusting the flow direction conditioned on new inputs or jumping in the vector field (see Section~\ref{temporal updates}), the evolving dynamics of graph structures have received limited attention. Most existing methods either generate graphs using SDEs in a discrete manner~\cite{liang2024dynamic} or adopt an end-to-end approach for graph generation tasks~\cite{verma_modular_nodate, huang_graphgdp_2022, huang2023conditional}. Although Huang et al. propose a framework that allows the dynamic evolution of both edge weights and node features, it overlooks the newly observed graph structures that emerge dynamically. Since graph topology significantly influences the evolution of node embeddings, it is crucial to incorporate the dynamic evolution of graph structures during inference to improve downstream performance.

\subsection{Modeling Hierarchical Graph Dynamics}

Hierarchical or Multi-scale data is essential for capturing complex structures and long-range dependencies that arise at different levels of granularity. While extensive research has been conducted on developing multi-scale GNNs~\cite{ying2018hierarchical, qiu2024msgnn}, relatively few studies have explored their integration with neural differential equations. From a data perspective, multi-scale characteristics can manifest at both temporal and spatial levels. From a modeling perspective, multi-scale information can be incorporated at different stages, such as during encoding or the modeling of differential equations, which yields multiple latent trajectories at different levels. Notably, Wang et al.~\cite{wang2024dynamic} capture multi-scale temporal information at the encoding stage before applying a graph-based ODE. However, the incorporation of multi-scale spatial modeling within the differential equation remains an open area of research under the framework of Graph NDEs.




\section{Conclusion}


In this survey, we presentes the first comprehensive overview of Graph Neural Differential Equations (Graph NDEs), beginning with the fundamental concepts of both GNNs and differential equations. We then introduce a structured taxonomy covering tasks, graph construction methods, and the roles of GNNs in various settings. Methodologically, we analyze existing literature through two primary perspectives: Temporal Dynamics Modeling and Spatial Dynamics Modeling, highlighting key challenges and potential solutions. Additionally, we discuss diverse applications of Graph NDEs and identify persistent research gaps, suggesting directions for future study. By detailing how GNNs can be integrated more naturally with differential equation frameworks, we believe this survey will serve as a catalyst for continued innovation in this rapidly developing field, inspiring both researchers and practitioners to advance the state of the art in Graph NDEs.

\section*{Acknowledgment}
This research was partially supported by the US National Science Foundation under Award Number 2319449 and Award Number 2312502, as well as the US National Institute of Diabetes and Digestive and Kidney Diseases of the US National Institutes of Health under Award Number K25DK135913.

\bibliography{ref}

\begin{thebibliography}{100}

\bibitem{keane2017climate}
Andrew Keane, Bernd Krauskopf, and Claire~M Postlethwaite.
\newblock Climate models with delay differential equations.
\newblock {\em Chaos: An Interdisciplinary Journal of Nonlinear Science}, 27(11), 2017.

\bibitem{holmes1994partial}
Elizabeth~E Holmes, Mark~A Lewis, JE~Banks, and RR~Veit.
\newblock Partial differential equations in ecology: spatial interactions and population dynamics.
\newblock {\em Ecology}, 75(1):17--29, 1994.

\bibitem{yang2020parameter}
Xiangfeng Yang, Yuhan Liu, and Gyei-Kark Park.
\newblock Parameter estimation of uncertain differential equation with application to financial market.
\newblock {\em Chaos, Solitons \& Fractals}, 139:110026, 2020.

\bibitem{dang2023conditional}
Ting Dang, Jing Han, Tong Xia, Erika Bondareva, Chlo{\"e} Siegele-Brown, Jagmohan Chauhan, Andreas Grammenos, Dimitris Spathis, Pietro Cicuta, and Cecilia Mascolo.
\newblock Conditional neural ode processes for individual disease progression forecasting: a case study on covid-19.
\newblock In {\em Proceedings of the 29th ACM SIGKDD Conference On Knowledge Discovery and Data Mining}, pages 3914--3925, 2023.

\bibitem{wang2025ncode}
Xiaoda Wang, Yuji Zhao, Kaiqiao Han, Xiao Luo, Sanne van Rooij, Jennifer Stevens, Lifang He, Liang Zhan, Yizhou Sun, Wei Wang, and Carl Yang.
\newblock Conditional neural ode for longitudinal parkinson's disease progression forecasting.
\newblock In {\em Abstract in the Organization for Human Brain Mapping Annual Meeting}, 2025.

\bibitem{maki2013infectious}
Yoshihiro Maki and Hideo Hirose.
\newblock Infectious disease spread analysis using stochastic differential equations for sir model.
\newblock In {\em 2013 4th International Conference on Intelligent Systems, Modelling and Simulation}, pages 152--156. IEEE, 2013.

\bibitem{liu2024review}
Zewen Liu, Guancheng Wan, B~Aditya Prakash, Max~SY Lau, and Wei Jin.
\newblock A review of graph neural networks in epidemic modeling.
\newblock In {\em Proceedings of the 30th ACM SIGKDD Conference on Knowledge Discovery and Data Mining}, pages 6577--6587, 2024.

\bibitem{butcher2000numerical}
John~C Butcher.
\newblock Numerical methods for ordinary differential equations in the 20th century.
\newblock {\em Journal of Computational and Applied Mathematics}, 125(1-2):1--29, 2000.

\bibitem{sloan2012partial}
D~Sloan, S~Vandewalle, and E~S{\"u}li.
\newblock Partial differential equations.
\newblock 2012.

\bibitem{van1976stochastic}
Nicolaas~G Van~Kampen.
\newblock Stochastic differential equations.
\newblock {\em Physics reports}, 24(3):171--228, 1976.

\bibitem{todorovski2007integrating}
Ljup{\v{c}}o Todorovski and Sa{\v{s}}o D{\v{z}}eroski.
\newblock Integrating domain knowledge in equation discovery.
\newblock In {\em Computational Discovery of Scientific Knowledge: Introduction, Techniques, and Applications in Environmental and Life Sciences}, pages 69--97. Springer, 2007.

\bibitem{bryutkin_hamlet_2024}
Andrey Bryutkin, Jiahao Huang, Zhongying Deng, Guang Yang, Carola-Bibiane Schönlieb, and Angelica Aviles-Rivero.
\newblock {HAMLET}: {Graph} {Transformer} {Neural} {Operator} for {Partial} {Differential} {Equations}, October 2024.
\newblock arXiv:2402.03541 [cs].

\bibitem{kumar_grade_2021}
Yash Kumar and Souvik Chakraborty.
\newblock {GrADE}: {A} graph based data-driven solver for time-dependent nonlinear partial differential equations, August 2021.
\newblock arXiv:2108.10639 [stat].

\bibitem{choi_gnrk_2023}
Hoyun Choi, Sungyeop Lee, B.~Kahng, and Junghyo Jo.
\newblock {GNRK}: {Graph} {Neural} {Runge}-{Kutta} method for solving partial differential equations, October 2023.
\newblock arXiv:2310.00618 [cs].

\bibitem{chen2018neural}
Ricky~TQ Chen, Yulia Rubanova, Jesse Bettencourt, and David~K Duvenaud.
\newblock Neural ordinary differential equations.
\newblock {\em Advances in neural information processing systems}, 31, 2018.

\bibitem{wang2021literature}
Xiaomei Wang, Qi~An, Zilong He, and Wei Fang.
\newblock A literature review of social network analysis in epidemic prevention and control.
\newblock {\em Complexity}, 2021(1):3816221, 2021.

\bibitem{medina2022urban}
Boris Medina-Salgado, Eddy S{\'a}nchez-DelaCruz, Pilar Pozos-Parra, and Javier~E Sierra.
\newblock Urban traffic flow prediction techniques: A review.
\newblock {\em Sustainable Computing: Informatics and Systems}, 35:100739, 2022.

\bibitem{Wan_GraphODE_ICML25}
Guancheng Wan, Zijie Huang, Wanjia Zhao, Xiao Luo, Yizhou Sun, and Wei Wang.
\newblock Rethink graphode generalization within coupled dynamical system.
\newblock In {\em Forty-second International Conference on Machine Learning}, 2025.

\bibitem{scarselli2008graph}
Franco Scarselli, Marco Gori, Ah~Chung Tsoi, Markus Hagenbuchner, and Gabriele Monfardini.
\newblock The graph neural network model.
\newblock {\em IEEE transactions on neural networks}, 20(1):61--80, 2008.

\bibitem{wu2019simplifying}
Felix Wu, Amauri Souza, Tianyi Zhang, Christopher Fifty, Tao Yu, and Kilian Weinberger.
\newblock Simplifying graph convolutional networks.
\newblock In {\em International conference on machine learning}, pages 6861--6871. Pmlr, 2019.

\bibitem{liu2024tinygraph}
Yezi Liu and Yanning Shen.
\newblock Tinygraph: joint feature and node condensation for graph neural networks.
\newblock {\em arXiv preprint arXiv:2407.08064}, 2024.

\bibitem{poli2019graph}
Michael Poli, Stefano Massaroli, Junyoung Park, Atsushi Yamashita, Hajime Asama, and Jinkyoo Park.
\newblock Graph neural ordinary differential equations.
\newblock {\em arXiv preprint arXiv:1911.07532}, 2019.

\bibitem{xhonneux2020continuous}
Louis-Pascal Xhonneux, Meng Qu, and Jian Tang.
\newblock Continuous graph neural networks.
\newblock In {\em International conference on machine learning}, pages 10432--10441. PMLR, 2020.

\bibitem{tang2024interpretable}
Haoteng Tang, Guodong Liu, Siyuan Dai, Kai Ye, Kun Zhao, Wenlu Wang, Carl Yang, Lifang He, Alex Leow, Paul Thompson, et~al.
\newblock Interpretable spatio-temporal embedding for brain structural-effective network with ordinary differential equation.
\newblock In {\em International Conference on Medical Image Computing and Computer-Assisted Intervention}, pages 227--237. Springer, 2024.

\bibitem{bryutkin2024hamlet}
Andrey Bryutkin, Jiahao Huang, Zhongying Deng, Guang Yang, Carola-Bibiane Sch{\"o}nlieb, and Angelica Aviles-Rivero.
\newblock Hamlet: Graph transformer neural operator for partial differential equations.
\newblock {\em arXiv preprint arXiv:2402.03541}, 2024.

\bibitem{bishnoi2024brognet}
Suresh Bishnoi, Jayadeva Jayadeva, Sayan Ranu, and NM~Anoop Krishnan.
\newblock Brognet: Momentum-conserving graph neural stochastic differential equation for learning brownian dynamics.
\newblock In {\em The Twelfth International Conference on Learning Representations}, 2024.

\bibitem{niu2024applications}
Hao Niu, Yuxiang Zhou, Xiaohao Yan, Jun Wu, Yuncheng Shen, Zhang Yi, and Junjie Hu.
\newblock On the applications of neural ordinary differential equations in medical image analysis.
\newblock {\em Artificial Intelligence Review}, 57(9):236, 2024.

\bibitem{losada2024bridging}
Idris~Bachali Losada and Nadia Terranova.
\newblock Bridging pharmacology and neural networks: A deep dive into neural ordinary differential equations.
\newblock {\em CPT: Pharmacometrics \& Systems Pharmacology}, 13(8):1289--1296, 2024.

\bibitem{dubeydeep}
Sourabh~Kumar Dubey, Hibah Islahi, and Raghvendra Singh.
\newblock Deep neural networks for solving ordinary differential equations: A comprehensive review.

\bibitem{han2023continuous}
Andi Han, Dai Shi, Lequan Lin, and Junbin Gao.
\newblock From continuous dynamics to graph neural networks: Neural diffusion and beyond.
\newblock {\em arXiv preprint arXiv:2310.10121}, 2023.

\bibitem{soleymani2024structure}
Farzan Soleymani, Eric Paquet, Herna~Lydia Viktor, and Wojtek Michalowski.
\newblock Structure-based protein and small molecule generation using egnn and diffusion models: A comprehensive review.
\newblock {\em Computational and Structural Biotechnology Journal}, 2024.

\bibitem{xie2022semisupervised}
Yu~Xie, Yanfeng Liang, Maoguo Gong, A~Kai Qin, Yew-Soon Ong, and Tiantian He.
\newblock Semisupervised graph neural networks for graph classification.
\newblock {\em IEEE Transactions on Cybernetics}, 53(10):6222--6235, 2022.

\bibitem{belbute2020combining}
Filipe De~Avila Belbute-Peres, Thomas Economon, and Zico Kolter.
\newblock Combining differentiable pde solvers and graph neural networks for fluid flow prediction.
\newblock In {\em international conference on machine learning}, pages 2402--2411. PMLR, 2020.

\bibitem{bergna_uncertainty_2024}
Richard Bergna, Sergio Calvo-Ordoñez, Felix~L. Opolka, Pietro Liò, and Jose~Miguel Hernandez-Lobato.
\newblock Uncertainty {Modeling} in {Graph} {Neural} {Networks} via {Stochastic} {Differential} {Equations}, September 2024.
\newblock arXiv:2408.16115 [cs].

\bibitem{liu2023group}
Shengchao Liu, Weitao Du, Zhi-Ming Ma, Hongyu Guo, and Jian Tang.
\newblock A group symmetric stochastic differential equation model for molecule multi-modal pretraining.
\newblock In {\em International Conference on Machine Learning}, pages 21497--21526. PMLR, 2023.

\bibitem{pontryagin2018mathematical}
Lev~Semenovich Pontryagin.
\newblock {\em Mathematical theory of optimal processes}.
\newblock Routledge, 2018.

\bibitem{zhang_improving_2022}
Yanfu Zhang, Shangqian Gao, Jian Pei, and Heng Huang.
\newblock Improving {Social} {Network} {Embedding} via {New} {Second}-{Order} {Continuous} {Graph} {Neural} {Networks}.
\newblock In {\em Proceedings of the 28th {ACM} {SIGKDD} {Conference} on {Knowledge} {Discovery} and {Data} {Mining}}, pages 2515--2523, Washington DC USA, August 2022. ACM.

\bibitem{huang2021str}
Chuyu Huang.
\newblock Str-godes: Spatial-temporal-ridership graph odes for metro ridership prediction.
\newblock {\em arXiv preprint arXiv:2107.04980}, 2021.

\bibitem{guo_evolutionary_2022}
Jiayan Guo, Peiyan Zhang, Chaozhuo Li, Xing Xie, Yan Zhang, and Sunghun Kim.
\newblock Evolutionary {Preference} {Learning} via {Graph} {Nested} {GRU} {ODE} for {Session}-based {Recommendation}.
\newblock In {\em Proceedings of the 31st {ACM} {International} {Conference} on {Information} \& {Knowledge} {Management}}, pages 624--634, October 2022.
\newblock arXiv:2206.12779 [cs].

\bibitem{wang_causalgnn_2022}
Lijing Wang, Aniruddha Adiga, Jiangzhuo Chen, Adam Sadilek, Srinivasan Venkatramanan, and Madhav Marathe.
\newblock {CausalGNN}: {Causal}-{Based} {Graph} {Neural} {Networks} for {Spatio}-{Temporal} {Epidemic} {Forecasting}.
\newblock {\em Proceedings of the AAAI Conference on Artificial Intelligence}, 36(11):12191--12199, June 2022.

\bibitem{huang2021coupled}
Zijie Huang, Yizhou Sun, and Wei Wang.
\newblock Coupled graph ode for learning interacting system dynamics.
\newblock In {\em Proceedings of the 27th ACM SIGKDD conference on knowledge discovery \& data mining}, pages 705--715, 2021.

\bibitem{huang_tango_2023}
Zijie Huang, Wanjia Zhao, Jingdong Gao, Ziniu Hu, Xiao Luo, Yadi Cao, Yuanzhou Chen, Yizhou Sun, and Wei Wang.
\newblock {TANGO}: {Time}-{Reversal} {Latent} {GraphODE} for {Multi}-{Agent} {Dynamical} {Systems}, October 2023.
\newblock arXiv:2310.06427 [cs].

\bibitem{wang_adding_2024}
Peixiao Wang, Tong Zhang, Hengcai Zhang, Shifen Cheng, and Wangshu Wang.
\newblock Adding attention to the neural ordinary differential equation for spatio-temporal prediction.
\newblock {\em International Journal of Geographical Information Science}, 38(1):156--181, January 2024.

\bibitem{xunode}
Ke~Xu, Weizhi Zhang, Yuanjie Zhu, Zihe Song, and S~Yu Philip.
\newblock Node-sat: Temporal graph learning with neural ode-guided self-attention.

\bibitem{xiong_diffusion_2023}
Ni~Xiong, Yan Yang, Yongquan Jiang, and Xiaocao Ouyang.
\newblock Diffusion {Graph} {Neural} {Ordinary} {Differential} {Equation} {Network} for {Traffic} {Prediction}.
\newblock In {\em 2023 {International} {Joint} {Conference} on {Neural} {Networks} ({IJCNN})}, pages 1--8, June 2023.
\newblock ISSN: 2161-4407.

\bibitem{yan_contig_2021}
Xu~Yan, Xiaoliang Fan, Peizhen Yang, Zonghan Wu, Shirui Pan, Longbiao Chen, Yu~Zang, and Cheng Wang.
\newblock {ConTIG}: {Continuous} {Representation} {Learning} on {Temporal} {Interaction} {Graphs}, September 2021.
\newblock arXiv:2110.06088 [cs].

\bibitem{zang_neural_2020}
Chengxi Zang and Fei Wang.
\newblock Neural {Dynamics} on {Complex} {Networks}.
\newblock In {\em Proceedings of the 26th {ACM} {SIGKDD} {International} {Conference} on {Knowledge} {Discovery} \& {Data} {Mining}}, pages 892--902, Virtual Event CA USA, August 2020. ACM.

\bibitem{huang2020learning}
Zijie Huang, Yizhou Sun, and Wei Wang.
\newblock Learning continuous system dynamics from irregularly-sampled partial observations.
\newblock {\em Advances in Neural Information Processing Systems}, 33:16177--16187, 2020.

\bibitem{desai_variational_2021}
Shaan Desai, Marios Mattheakis, and Stephen Roberts.
\newblock Variational {Integrator} {Graph} {Networks} for {Learning} {Energy} {Conserving} {Dynamical} {Systems}.
\newblock {\em Physical Review E}, 104(3):035310, September 2021.
\newblock arXiv:2004.13688 [cs].

\bibitem{fang_spatial-temporal_2021}
Zheng Fang, Qingqing Long, Guojie Song, and Kunqing Xie.
\newblock Spatial-{Temporal} {Graph} {ODE} {Networks} for {Traffic} {Flow} {Forecasting}.
\newblock In {\em Proceedings of the 27th {ACM} {SIGKDD} {Conference} on {Knowledge} {Discovery} \& {Data} {Mining}}, pages 364--373, August 2021.
\newblock arXiv:2106.12931 [cs].

\bibitem{iakovlev_learning_2021}
Valerii Iakovlev, Markus Heinonen, and Harri Lähdesmäki.
\newblock Learning continuous-time {PDEs} from sparse data with graph neural networks, January 2021.
\newblock arXiv:2006.08956 [cs].

\bibitem{rusch_graph-coupled_2022}
T.~Konstantin Rusch, Ben Chamberlain, James Rowbottom, Siddhartha Mishra, and Michael Bronstein.
\newblock Graph-{Coupled} {Oscillator} {Networks}.
\newblock In {\em Proceedings of the 39th {International} {Conference} on {Machine} {Learning}}, pages 18888--18909. PMLR, June 2022.
\newblock ISSN: 2640-3498.

\bibitem{luo_hope_2023}
Xiao Luo, Jingyang Yuan, Zijie Huang, Huiyu Jiang, Yifang Qin, Wei Ju, Ming Zhang, and Yizhou Sun.
\newblock {HOPE}: {High}-order {Graph} {ODE} {For} {Modeling} {Interacting} {Dynamics}.
\newblock In {\em Proceedings of the 40th {International} {Conference} on {Machine} {Learning}}, pages 23124--23139. PMLR, July 2023.
\newblock ISSN: 2640-3498.

\bibitem{niwa_coordinet_2023}
Kenta Niwa, Naonori Ueda, Hiroshi Sawada, Akinori Fujino, Shoichiro Takeda, Guoqiang Zhang, and W.~Bastiaan Kleijn.
\newblock {CoordiNet}: {Constrained} {Dynamics} {Learning} for {State} {Coordination} {Over} {Graph}.
\newblock {\em IEEE Transactions on Signal and Information Processing over Networks}, 9:242--257, 2023.
\newblock Conference Name: IEEE Transactions on Signal and Information Processing over Networks.

\bibitem{wen_second-order_2024}
Song Wen, Hao Wang, Di~Liu, Qilong Zhangli, and Dimitris Metaxas.
\newblock Second-{Order} {Graph} {ODEs} for {Multi}-{Agent} {Trajectory} {Forecasting}.
\newblock In {\em 2024 {IEEE}/{CVF} {Winter} {Conference} on {Applications} of {Computer} {Vision} ({WACV})}, pages 5079--5088, Waikoloa, HI, USA, January 2024. IEEE.

\bibitem{liu_segno_2024}
Yang Liu, Jiashun Cheng, Haihong Zhao, Tingyang Xu, Peilin Zhao, Fugee Tsung, Jia Li, and Yu~Rong.
\newblock {SEGNO}: {Generalizing} {Equivariant} {Graph} {Neural} {Networks} with {Physical} {Inductive} {Biases}, March 2024.
\newblock arXiv:2308.13212 [cs].

\bibitem{eliasof_temporal_2024}
Moshe Eliasof, Eldad Haber, Eran Treister, and Carola-Bibiane~B. Schönlieb.
\newblock On {The} {Temporal} {Domain} of {Differential} {Equation} {Inspired} {Graph} {Neural} {Networks}.
\newblock In {\em Proceedings of {The} 27th {International} {Conference} on {Artificial} {Intelligence} and {Statistics}}, pages 1792--1800. PMLR, April 2024.
\newblock ISSN: 2640-3498.

\bibitem{huang_link_2024}
Liyi Huang, Bowen Pang, Qiming Yang, Xiangnan Feng, and Wei Wei.
\newblock Link prediction by continuous spatiotemporal representation via neural differential equations.
\newblock {\em Knowledge-Based Systems}, 292:111619, May 2024.

\bibitem{gao_egpde-net_2023}
Penglei Gao, Xi~Yang, Rui Zhang, Ping Guo, John~Y. Goulermas, and Kaizhu Huang.
\newblock {EgPDE}-{Net}: {Building} {Continuous} {Neural} {Networks} for {Time} {Series} {Prediction} with {Exogenous} {Variables}, September 2023.
\newblock arXiv:2208.01913 [cs].

\bibitem{kosma_neural_nodate}
Chrysoula Kosma, Giannis Nikolentzos, George Panagopoulos, Jean-Marc Steyaert, and Michalis Vazirgiannis.
\newblock Neural {Ordinary} {Differential} {Equations} for {Modeling} {Epidemic} {Spreading}.

\bibitem{uwamichi_integrating_2024}
Masahito Uwamichi, Simon~K. Schnyder, Tetsuya~J. Kobayashi, and Satoshi Sawai.
\newblock Integrating {GNN} and {Neural} {ODEs} for {Estimating} {Non}-{Reciprocal} {Two}-{Body} {Interactions} in {Mixed}-{Species} {Collective} {Motion}, November 2024.
\newblock arXiv:2405.16503 [physics].

\bibitem{yuan2024egode}
Jingyang Yuan, Gongbo Sun, Zhiping Xiao, Hang Zhou, Xiao Luo, Junyu Luo, Yusheng Zhao, Wei Ju, and Ming Zhang.
\newblock Egode: An event-attended graph ode framework for modeling rigid dynamics.
\newblock In {\em The Thirty-eighth Annual Conference on Neural Information Processing Systems}, 2024.

\bibitem{choi2021lt}
Jeongwhan Choi, Jinsung Jeon, and Noseong Park.
\newblock Lt-ocf: Learnable-time ode-based collaborative filtering.
\newblock In {\em Proceedings of the 30th ACM International Conference on Information \& Knowledge Management}, pages 251--260, 2021.

\bibitem{yang2024siamese}
Yuxuan Yang, Siyuan Zhou, He~Weng, Dongjing Wang, Xin Zhang, Dongjin Yu, and Shuiguang Deng.
\newblock Siamese learning based on graph differential equation for next-poi recommendation.
\newblock {\em Applied Soft Computing}, 150:111086, 2024.

\bibitem{han2021learning}
Zhen Han, Zifeng Ding, Yunpu Ma, Yujia Gu, and Volker Tresp.
\newblock Learning neural ordinary equations for forecasting future links on temporal knowledge graphs.
\newblock In {\em Proceedings of the 2021 conference on empirical methods in natural language processing}, pages 8352--8364, 2021.

\bibitem{wang_acmp_2023}
Yuelin Wang, Kai Yi, Xinliang Liu, Yu~Guang Wang, and Shi Jin.
\newblock {ACMP}: {Allen}-{Cahn} {Message} {Passing} for {Graph} {Neural} {Networks} with {Particle} {Phase} {Transition}, April 2023.
\newblock arXiv:2206.05437 [cs].

\bibitem{han_learning_2021}
Zhen Han, Zifeng Ding, Yunpu Ma, Yujia Gu, and Volker Tresp.
\newblock Learning {Neural} {Ordinary} {Equations} for {Forecasting} {Future} {Links} on {Temporal} {Knowledge} {Graphs}.
\newblock In Marie-Francine Moens, Xuanjing Huang, Lucia Specia, and Scott Wen-tau Yih, editors, {\em Proceedings of the 2021 {Conference} on {Empirical} {Methods} in {Natural} {Language} {Processing}}, pages 8352--8364, Online and Punta Cana, Dominican Republic, November 2021. Association for Computational Linguistics.

\bibitem{wu_continuously_2024}
Jiajia Wu and Ling Chen.
\newblock Continuously {Evolving} {Graph} {Neural} {Controlled} {Differential} {Equations} for {Traffic} {Forecasting}, January 2024.
\newblock arXiv:2401.14695 [cs].

\bibitem{ma_spatio-temporal_2024}
Wenming Ma, Zihao Chu, Hao Chen, and Mingqi Li.
\newblock Spatio-temporal envolutional graph neural network for traffic flow prediction in {UAV}-based urban traffic monitoring system.
\newblock {\em Scientific Reports}, 14(1):26800, November 2024.

\bibitem{zhong_attention-based_2023}
Weiheng Zhong, Hadi Meidani, and Jane Macfarlane.
\newblock Attention-based {Spatial}-{Temporal} {Graph} {Neural} {ODE} for {Traffic} {Prediction}, May 2023.
\newblock arXiv:2305.00985 [cs].

\bibitem{noauthor_conditional_nodate}
Conditional {Diffusion} {Based} on {Discrete} {Graph} {Structures} for {Molecular} {Graph} {Generation} {\textbar} {Proceedings} of the {AAAI} {Conference} on {Artificial} {Intelligence}.

\bibitem{huang_graphgdp_2022}
Han Huang, Leilei Sun, Bowen Du, Yanjie Fu, and Weifeng Lv.
\newblock {GraphGDP}: {Generative} {Diffusion} {Processes} for {Permutation} {Invariant} {Graph} {Generation}, December 2022.
\newblock arXiv:2212.01842 [cs].

\bibitem{klemmer2023positional}
Konstantin Klemmer, Nathan~S Safir, and Daniel~B Neill.
\newblock Positional encoder graph neural networks for geographic data.
\newblock In {\em International conference on artificial intelligence and statistics}, pages 1379--1389. PMLR, 2023.

\bibitem{ma2023language}
Wenxuan Ma, Shuang Li, Jingxuan Kang, et~al.
\newblock Language semantic graph guided data-efficient learning.
\newblock {\em Advances in Neural Information Processing Systems}, 36:24088--24102, 2023.

\bibitem{jiahao_learning_2021}
Tom~Z. Jiahao, Lishuo Pan, and M.~Ani Hsieh.
\newblock Learning to {Swarm} with {Knowledge}-{Based} {Neural} {Ordinary} {Differential} {Equations}, December 2021.
\newblock arXiv:2109.04927 [cs].

\bibitem{chamberlain_grand_2021}
Ben Chamberlain, James Rowbottom, Maria~I. Gorinova, Michael Bronstein, Stefan Webb, and Emanuele Rossi.
\newblock {GRAND}: {Graph} {Neural} {Diffusion}.
\newblock In {\em Proceedings of the 38th {International} {Conference} on {Machine} {Learning}}, pages 1407--1418. PMLR, July 2021.
\newblock ISSN: 2640-3498.

\bibitem{chamberlain_beltrami_2021}
Benjamin Chamberlain, James Rowbottom, Davide Eynard, Francesco Di~Giovanni, Xiaowen Dong, and Michael Bronstein.
\newblock Beltrami {Flow} and {Neural} {Diffusion} on {Graphs}.
\newblock In {\em Advances in {Neural} {Information} {Processing} {Systems}}, volume~34, pages 1594--1609. Curran Associates, Inc., 2021.

\bibitem{choi_gread_2023}
Jeongwhan Choi, Seoyoung Hong, Noseong Park, and Sung-Bae Cho.
\newblock {GREAD}: {Graph} {Neural} {Reaction}-{Diffusion} {Networks}.
\newblock In {\em Proceedings of the 40th {International} {Conference} on {Machine} {Learning}}, pages 5722--5747. PMLR, July 2023.
\newblock ISSN: 2640-3498.

\bibitem{zhang_cope_2021}
Yao Zhang, Yun Xiong, Dongsheng Li, Caihua Shan, Kan Ren, and Yangyong Zhu.
\newblock {CoPE}: {Modeling} {Continuous} {Propagation} and {Evolution} on {Interaction} {Graph}.
\newblock In {\em Proceedings of the 30th {ACM} {International} {Conference} on {Information} \& {Knowledge} {Management}}, {CIKM} '21, pages 2627--2636, New York, NY, USA, 2021. Association for Computing Machinery.

\bibitem{xing_aggdn_2024}
Yucheng Xing, Jacqueline Wu, Yingru Liu, Xuewen Yang, and Xin Wang.
\newblock {AGGDN}: {A} {Continuous} {Stochastic} {Predictive} {Model} for {Monitoring} {Sporadic} {Time} {Series} on {Graphs}.
\newblock In Biao Luo, Long Cheng, Zheng-Guang Wu, Hongyi Li, and Chaojie Li, editors, {\em Neural {Information} {Processing}}, pages 130--146, Singapore, 2024. Springer Nature.

\bibitem{jin_multivariate_2023}
Ming Jin, Yu~Zheng, Yuan-Fang Li, Siheng Chen, Bin Yang, and Shirui Pan.
\newblock Multivariate {Time} {Series} {Forecasting} with {Dynamic} {Graph} {Neural} {ODEs}.
\newblock {\em IEEE Transactions on Knowledge and Data Engineering}, 35(9):9168--9180, September 2023.
\newblock arXiv:2202.08408 [cs].

\bibitem{Huang2020LGODE}
Zijie Huang, Yizhou Sun, and Wei Wang.
\newblock Learning continuous system dynamics from irregularly-sampled partial observations.
\newblock In {\em Advances in Neural Information Processing Systems 33 (NeurIPS 2020)}, 2020.

\bibitem{Huang2021CGODE}
Zijie Huang, Yizhou Sun, and Wei Wang.
\newblock Coupled graph ode for learning interacting system dynamics.
\newblock In {\em Proceedings of the 27th ACM SIGKDD Conference on Knowledge Discovery and Data Mining (KDD '21)}, pages 705--715, 2021.

\bibitem{Huang2023GGODE}
Zijie Huang, Yizhou Sun, and Wei Wang.
\newblock Generalizing graph ode for learning complex system dynamics across environments.
\newblock In {\em Proceedings of the 29th ACM SIGKDD Conference on Knowledge Discovery and Data Mining (KDD '23)}, 2023.

\bibitem{long2024unveiling}
Qingqing Long, Zheng Fang, Chen Fang, Chong Chen, Pengfei Wang, and Yuanchun Zhou.
\newblock Unveiling delay effects in traffic forecasting: a perspective from spatial-temporal delay differential equations.
\newblock In {\em Proceedings of the ACM Web Conference 2024}, pages 1035--1044, 2024.

\bibitem{li_physics-informed_2024}
Jiahao Li, Huandong Wang, and Xinlei Chen.
\newblock Physics-informed {Neural} {ODE} for {Post}-disaster {Mobility} {Recovery}.
\newblock In {\em Proceedings of the 30th {ACM} {SIGKDD} {Conference} on {Knowledge} {Discovery} and {Data} {Mining}}, pages 1587--1598, Barcelona Spain, August 2024. ACM.

\bibitem{han2023devil}
Zhenyu Han, Yanxin Xi, Tong Xia, Yu~Liu, and Yong Li.
\newblock Devil in the landscapes: Inferring epidemic exposure risks from street view imagery.
\newblock In {\em Proceedings of the 31st ACM International Conference on Advances in Geographic Information Systems}, pages 1--4, 2023.

\bibitem{sanchez2019hamiltonian}
Alvaro Sanchez-Gonzalez, Victor Bapst, Kyle Cranmer, and Peter Battaglia.
\newblock Hamiltonian graph networks with ode integrators.
\newblock {\em arXiv preprint arXiv:1909.12790}, 2019.

\bibitem{cranmer2020lagrangian}
Miles Cranmer, Sam Greydanus, Stephan Hoyer, Peter Battaglia, David Spergel, and Shirley Ho.
\newblock Lagrangian neural networks.
\newblock {\em arXiv preprint arXiv:2003.04630}, 2020.

\bibitem{cecil1977finite}
Zienkiewicz~Olgierd Cecil, Taylor~Robert Leroy, Nithiarasu Perumal, and Zhu JZ.
\newblock The finite element method. vol. 3, 1977.

\bibitem{jiang_neural_2023}
Zichao Jiang, Junyang Jiang, Qinghe Yao, and Gengchao Yang.
\newblock A neural network-based {PDE} solving algorithm with high precision.
\newblock {\em Scientific Reports}, 13(1):4479, March 2023.

\bibitem{han2013numerical}
Tianmin Han and Yuhuan Han.
\newblock Numerical solution for super large scale systems.
\newblock {\em IEEE Access}, 1:537--544, 2013.

\bibitem{rusch2023survey}
T~Konstantin Rusch, Michael~M Bronstein, and Siddhartha Mishra.
\newblock A survey on oversmoothing in graph neural networks.
\newblock {\em arXiv preprint arXiv:2303.10993}, 2023.

\bibitem{he2016deep}
Kaiming He, Xiangyu Zhang, Shaoqing Ren, and Jian Sun.
\newblock Deep residual learning for image recognition.
\newblock In {\em Proceedings of the IEEE conference on computer vision and pattern recognition}, pages 770--778, 2016.

\bibitem{xu2021optimization}
Keyulu Xu, Mozhi Zhang, Stefanie Jegelka, and Kenji Kawaguchi.
\newblock Optimization of graph neural networks: Implicit acceleration by skip connections and more depth.
\newblock In {\em International Conference on Machine Learning}, pages 11592--11602. PMLR, 2021.

\bibitem{finlay2020train}
Chris Finlay, J{\"o}rn-Henrik Jacobsen, Levon Nurbekyan, and Adam Oberman.
\newblock How to train your neural ode: the world of jacobian and kinetic regularization.
\newblock In {\em International conference on machine learning}, pages 3154--3164. PMLR, 2020.

\bibitem{thorpe_grand_2022}
Matthew Thorpe, Hedi Xia, Tan Nguyen, Thomas Strohmer, Andrea~L Bertozzi, and Stanley~J Osher.
\newblock {GRAND}++: {GRAPH} {NEURAL} {DIFFUSION} {WITH} {A} {SOURCE} {TERM}.
\newblock 2022.

\bibitem{maskey_fractional_nodate}
Sohir Maskey, Raffaele Paolino, Aras Bacho, and Gitta Kutyniok.
\newblock A {Fractional} {Graph} {Laplacian} {Approach} to {Oversmoothing}.

\bibitem{hao2021ks}
Yu~Hao, Xin Cao, Yufan Sheng, Yixiang Fang, and Wei Wang.
\newblock Ks-gnn: Keywords search over incomplete graphs via graphs neural network.
\newblock {\em Advances in Neural Information Processing Systems}, 34:1700--1712, 2021.

\bibitem{wu2024graph}
Qitian Wu, Fan Nie, Chenxiao Yang, Tianyi Bao, and Junchi Yan.
\newblock Graph out-of-distribution generalization via causal intervention.
\newblock In {\em Proceedings of the ACM Web Conference 2024}, pages 850--860, 2024.

\bibitem{bergna_graph_2023}
Richard Bergna, Felix Opolka, Pietro Liò, and Jose~Miguel Hernandez-Lobato.
\newblock Graph {Neural} {Stochastic} {Differential} {Equations}, August 2023.
\newblock arXiv:2308.12316 [cs].

\bibitem{lin2024graph}
Xixun Lin, Wenxiao Zhang, Fengzhao Shi, Chuan Zhou, Lixin Zou, Xiangyu Zhao, Dawei Yin, Shirui Pan, and Yanan Cao.
\newblock Graph neural stochastic diffusion for estimating uncertainty in node classification.
\newblock In {\em Forty-first International Conference on Machine Learning}, 2024.

\bibitem{liang2024dynamic}
Guojun Liang, Prayag Tiwari, S{\l}awomir Nowaczyk, Stefan Byttner, and Fernando Alonso-Fernandez.
\newblock Dynamic causal explanation based diffusion-variational graph neural network for spatiotemporal forecasting.
\newblock {\em IEEE Transactions on Neural Networks and Learning Systems}, 2024.

\bibitem{song2022robustness}
Yang Song, Qiyu Kang, Sijie Wang, Kai Zhao, and Wee~Peng Tay.
\newblock On the robustness of graph neural diffusion to topology perturbations.
\newblock {\em Advances in Neural Information Processing Systems}, 35:6384--6396, 2022.

\bibitem{zhao2024adversarial}
Kai Zhao, Qiyu Kang, Yang Song, Rui She, Sijie Wang, and Wee~Peng Tay.
\newblock Adversarial robustness in graph neural networks: A hamiltonian approach.
\newblock {\em Advances in Neural Information Processing Systems}, 36, 2024.

\bibitem{kang2024coupling}
Qiyu Kang, Kai Zhao, Yang Song, Yihang Xie, Yanan Zhao, Sijie Wang, Rui She, and Wee~Peng Tay.
\newblock Coupling graph neural networks with fractional order continuous dynamics: A robustness study.
\newblock In {\em Proceedings of the AAAI Conference on Artificial Intelligence}, volume~38, pages 13049--13058, 2024.

\bibitem{zhao2023graph}
Kai Zhao, Qiyu Kang, Yang Song, Rui She, Sijie Wang, and Wee~Peng Tay.
\newblock Graph neural convection-diffusion with heterophily.
\newblock {\em arXiv preprint arXiv:2305.16780}, 2023.

\bibitem{zhang2024unleashing}
Acong Zhang and Ping Li.
\newblock Unleashing the power of high-pass filtering in continuous graph neural networks.
\newblock In {\em Asian Conference on Machine Learning}, pages 1683--1698. PMLR, 2024.

\bibitem{feng2019hypergraph}
Yifan Feng, Haoxuan You, Zizhao Zhang, Rongrong Ji, and Yue Gao.
\newblock Hypergraph neural networks.
\newblock In {\em Proceedings of the AAAI conference on artificial intelligence}, volume~33, pages 3558--3565, 2019.

\bibitem{yao_spatio-temporal_2023}
Chengzhi Yao, Zhi Li, and Junbo Wang.
\newblock Spatio-{Temporal} {Hypergraph} {Neural} {ODE} {Network} for {Traffic} {Forecasting}.
\newblock In {\em 2023 {IEEE} {International} {Conference} on {Data} {Mining} ({ICDM})}, pages 1499--1504, December 2023.
\newblock ISSN: 2374-8486.

\bibitem{yan_hypergraph_2024}
Jielong Yan, Yifan Feng, Shihui Ying, and Yue Gao.
\newblock {HYPERGRAPH} {DYNAMIC} {SYSTEM}.
\newblock 2024.

\bibitem{shi2024towards}
Guangsi Shi, Daokun Zhang, Ming Jin, Shirui Pan, and S~Yu Philip.
\newblock Towards complex dynamic physics system simulation with graph neural ordinary equations.
\newblock {\em Neural Networks}, 176:106341, 2024.

\bibitem{huang2024physics}
Zijie Huang, Wanjia Zhao, Jingdong Gao, Ziniu Hu, Xiao Luo, Yadi Cao, Yuanzhou Chen, Yizhou Sun, and Wei Wang.
\newblock Physics-informed regularization for domain-agnostic dynamical system modeling.
\newblock {\em arXiv preprint arXiv:2410.06366}, 2024.

\bibitem{wu2025pure}
Hao Wu, Changhu Wang, Fan Xu, Jinbao Xue, Chong Chen, Xian-Sheng Hua, and Xiao Luo.
\newblock Pure: Prompt evolution with graph ode for out-of-distribution fluid dynamics modeling.
\newblock {\em Advances in Neural Information Processing Systems}, 37:104965--104994, 2025.

\bibitem{Sanchez-Gonzalez2019HGN}
Alvaro Sanchez-Gonzalez, Victor Bapst, Kyle Cranmer, and Peter Battaglia.
\newblock Hamiltonian graph networks with ode integrators.
\newblock In {\em NeurIPS}, 2019.

\bibitem{Shi2023GNSTODE}
Guangsi Shi et~al.
\newblock Towards complex dynamic physics system simulation with graph neural odes.
\newblock {\em arXiv preprint arXiv:2305.12334}, 2023.

\bibitem{fang2021stgode}
Zheng Fang et~al.
\newblock Spatial-temporal graph ode networks for traffic flow forecasting.
\newblock In {\em KDD}, 2021.

\bibitem{su2022gode_rnn}
Yuqiao Su et~al.
\newblock Graph ode recurrent neural networks for traffic flow forecasting.
\newblock In {\em ICECE}, 2022.

\bibitem{zhong2023astgode}
Weiheng Zhong et~al.
\newblock Attention-based spatial-temporal graph neural ode for traffic prediction.
\newblock {\em arXiv preprint arXiv:2305.00985}, 2023.

\bibitem{liu2023gramode}
Zibo Liu et~al.
\newblock Graph-based multi-ode neural networks for spatio-temporal traffic forecasting.
\newblock {\em TMLR}, 2023.

\bibitem{bai2024agode}
Lin Bai et~al.
\newblock Adaptive correlation graph neural ordinary differential equation for traffic flow forecasting.
\newblock {\em Engineering Letters}, 2024.

\bibitem{da2020recommendation}
Aminu Da’u and Naomie Salim.
\newblock Recommendation system based on deep learning methods: a systematic review and new directions.
\newblock {\em Artificial Intelligence Review}, 53(4):2709--2748, 2020.

\bibitem{su2009survey}
Xiaoyuan Su and Taghi~M Khoshgoftaar.
\newblock A survey of collaborative filtering techniques.
\newblock {\em Advances in artificial intelligence}, 2009(1):421425, 2009.

\bibitem{qin_learning_2024}
Yifang Qin, Wei Ju, Hongjun Wu, Xiao Luo, and Ming Zhang.
\newblock Learning {Graph} {ODE} for {Continuous}-{Time} {Sequential} {Recommendation}.
\newblock {\em IEEE Transactions on Knowledge and Data Engineering}, 36(7):3224--3236, July 2024.
\newblock arXiv:2304.07042 [cs].

\bibitem{yang_siamese_2024}
Yuxuan Yang, Siyuan Zhou, He~Weng, Dongjing Wang, Xin Zhang, Dongjin Yu, and Shuiguang Deng.
\newblock Siamese learning based on graph differential equation for {Next}-{POI} recommendation.
\newblock {\em Applied Soft Computing}, 150:111086, January 2024.

\bibitem{duan2015mathematical}
Wei Duan, Zongchen Fan, Peng Zhang, Gang Guo, and Xiaogang Qiu.
\newblock Mathematical and computational approaches to epidemic modeling: a comprehensive review.
\newblock {\em Frontiers of Computer Science}, 9:806--826, 2015.

\bibitem{liu2024epilearn}
Zewen Liu, Yunxiao Li, Mingyang Wei, Guancheng Wan, Max~S.Y. Lau, and Wei Jin.
\newblock Epilearn: A python library for machine learning in epidemic modeling.
\newblock In {\em epiDAMIK 2024: The 7th International Workshop on Epidemiology meets Data Mining and Knowledge Discovery at KDD 2024}, 2024.

\bibitem{rodriguez2022data}
Alexander Rodr{\'\i}guez, Harshavardhan Kamarthi, Pulak Agarwal, Javen Ho, Mira Patel, Suchet Sapre, and B~Aditya Prakash.
\newblock Data-centric epidemic forecasting: A survey.
\newblock {\em arXiv preprint arXiv:2207.09370}, 2022.

\bibitem{gao2021stan}
Junyi Gao, Rakshith Sharma, Cheng Qian, Lucas~M Glass, Jeffrey Spaeder, Justin Romberg, Jimeng Sun, and Cao Xiao.
\newblock Stan: spatio-temporal attention network for pandemic prediction using real-world evidence.
\newblock {\em Journal of the American Medical Informatics Association}, 28(4):733--743, 2021.

\bibitem{cao2022mepognn}
Qi~Cao, Renhe Jiang, Chuang Yang, Zipei Fan, Xuan Song, and Ryosuke Shibasaki.
\newblock Mepognn: Metapopulation epidemic forecasting with graph neural networks.
\newblock In {\em Joint European conference on machine learning and knowledge discovery in databases}, pages 453--468. Springer, 2022.

\bibitem{wan_epidemiology-aware_2024}
Guancheng Wan, Zewen Liu, Max S.~Y. Lau, B.~Aditya Prakash, and Wei Jin.
\newblock Epidemiology-{Aware} {Neural} {ODE} with {Continuous} {Disease} {Transmission} {Graph}, November 2024.
\newblock arXiv:2410.00049 [cs].

\bibitem{zhu2022survey}
Yanqiao Zhu, Yuanqi Du, Yinkai Wang, Yichen Xu, Jieyu Zhang, Qiang Liu, and Shu Wu.
\newblock A survey on deep graph generation: Methods and applications.
\newblock In {\em Learning on Graphs Conference}, pages 47--1. PMLR, 2022.

\bibitem{doersch2016tutorial}
Carl Doersch.
\newblock Tutorial on variational autoencoders.
\newblock {\em arXiv preprint arXiv:1606.05908}, 2016.

\bibitem{niu2020permutation}
Chenhao Niu, Yang Song, Jiaming Song, Shengjia Zhao, Aditya Grover, and Stefano Ermon.
\newblock Permutation invariant graph generation via score-based generative modeling.
\newblock In {\em International conference on artificial intelligence and statistics}, pages 4474--4484. PMLR, 2020.

\bibitem{jo2022score}
Jaehyeong Jo, Seul Lee, and Sung~Ju Hwang.
\newblock Score-based generative modeling of graphs via the system of stochastic differential equations.
\newblock In {\em International conference on machine learning}, pages 10362--10383. PMLR, 2022.

\bibitem{huang2023conditional}
Han Huang, Leilei Sun, Bowen Du, and Weifeng Lv.
\newblock Conditional diffusion based on discrete graph structures for molecular graph generation.
\newblock In {\em Proceedings of the AAAI Conference on Artificial Intelligence}, volume~37, pages 4302--4311, 2023.

\bibitem{angelis2023artificial}
Dimitrios Angelis, Filippos Sofos, and Theodoros~E Karakasidis.
\newblock Artificial intelligence in physical sciences: Symbolic regression trends and perspectives.
\newblock {\em Archives of Computational Methods in Engineering}, 30(6):3845--3865, 2023.

\bibitem{biggio2021neural}
Luca Biggio, Tommaso Bendinelli, Alexander Neitz, Aurelien Lucchi, and Giambattista Parascandolo.
\newblock Neural symbolic regression that scales.
\newblock In {\em International Conference on Machine Learning}, pages 936--945. Pmlr, 2021.

\bibitem{shojaee2024llm}
Parshin Shojaee, Kazem Meidani, Shashank Gupta, Amir~Barati Farimani, and Chandan~K Reddy.
\newblock Llm-sr: Scientific equation discovery via programming with large language models.
\newblock {\em arXiv preprint arXiv:2404.18400}, 2024.

\bibitem{yan2024enhanced}
Jianxiang Yan, Guanghui Song, Ying Li, Zhaoji Zhang, and Yuhao Chi.
\newblock Enhanced odma with channel code design and pattern collision resolution for unsourced multiple access.
\newblock In {\em 2024 IEEE International Symposium on Information Theory (ISIT)}, pages 3201--3206. IEEE, 2024.

\bibitem{luo2023graph}
Linhao Luo, Gholamreza Haffari, and Shirui Pan.
\newblock Graph sequential neural ode process for link prediction on dynamic and sparse graphs.
\newblock In {\em Proceedings of the sixteenth ACM international conference on web search and data mining}, pages 778--786, 2023.

\bibitem{chen2023label}
Zhikai Chen, Haitao Mao, Hongzhi Wen, Haoyu Han, Wei Jin, Haiyang Zhang, Hui Liu, and Jiliang Tang.
\newblock Label-free node classification on graphs with large language models (llms).
\newblock {\em arXiv preprint arXiv:2310.04668}, 2023.

\bibitem{hashemi2024comprehensive}
Mohammad Hashemi, Shengbo Gong, Juntong Ni, Wenqi Fan, B~Aditya Prakash, and Wei Jin.
\newblock A comprehensive survey on graph reduction: Sparsification, coarsening, and condensation.
\newblock {\em arXiv preprint arXiv:2402.03358}, 2024.

\bibitem{finzi2023stable}
Marc Finzi, Andres Potapczynski, Matthew Choptuik, and Andrew~Gordon Wilson.
\newblock A stable and scalable method for solving initial value pdes with neural networks.
\newblock In {\em 11th International Conference on Learning Representations, ICLR 2023}, 2023.

\bibitem{han2018solving}
Jiequn Han, Arnulf Jentzen, and Weinan E.
\newblock Solving high-dimensional partial differential equations using deep learning.
\newblock {\em Proceedings of the National Academy of Sciences}, 115(34):8505--8510, 2018.

\bibitem{samat2021gpu}
Alim Samat, Erzhu Li, Peijun Du, Sicong Liu, and Junshi Xia.
\newblock Gpu-accelerated catboost-forest for hyperspectral image classification via parallelized mrmr ensemble subspace feature selection.
\newblock {\em IEEE Journal of Selected Topics in Applied Earth Observations and Remote Sensing}, 14:3200--3214, 2021.

\bibitem{jagtap2020conservative}
Ameya~D Jagtap, Ehsan Kharazmi, and George~Em Karniadakis.
\newblock Conservative physics-informed neural networks on discrete domains for conservation laws: Applications to forward and inverse problems.
\newblock {\em Computer Methods in Applied Mechanics and Engineering}, 365:113028, 2020.

\bibitem{verma_modular_nodate}
Yogesh Verma, Samuel Kaski, Markus Heinonen, and Vikas Garg.
\newblock Modular {Flows}: {Differential} {Molecular} {Generation}.

\bibitem{ying2018hierarchical}
Zhitao Ying, Jiaxuan You, Christopher Morris, Xiang Ren, Will Hamilton, and Jure Leskovec.
\newblock Hierarchical graph representation learning with differentiable pooling.
\newblock {\em Advances in neural information processing systems}, 31, 2018.

\bibitem{qiu2024msgnn}
Mingjie Qiu, Zhiyi Tan, and Bing-Kun Bao.
\newblock Msgnn: Multi-scale spatio-temporal graph neural network for epidemic forecasting.
\newblock {\em Data Mining and Knowledge Discovery}, 38(4):2348--2376, 2024.

\bibitem{wang2024dynamic}
Keyi Wang, Jichao Zhan, Qinghua Si, Yueying Li, and Youyong Kong.
\newblock Dynamic multi-scale spatio-temporal graph ode for metro ridership prediction.
\newblock In {\em 2024 IEEE 7th Advanced Information Technology, Electronic and Automation Control Conference (IAEAC)}, volume~7, pages 1501--1509. IEEE, 2024.

\bibitem{huang2024embedding}
Huimin Huang, Luodi Xie, Mingzhe Liu, Jiajun Lin, and Hong Shen.
\newblock An embedding model for temporal knowledge graphs with long and irregular intervals.
\newblock {\em Knowledge-Based Systems}, 296:111893, 2024.

\bibitem{cui2024graph}
Kaiyuan Cui, Xinyan Wang, Zicheng Zhang, and Weichen Zhao.
\newblock Graph neural aggregation-diffusion with metastability.
\newblock {\em arXiv preprint arXiv:2403.20221}, 2024.

\bibitem{chamberlain2021grand}
Ben Chamberlain, James Rowbottom, Maria~I Gorinova, Michael Bronstein, Stefan Webb, and Emanuele Rossi.
\newblock Grand: Graph neural diffusion.
\newblock In {\em International conference on machine learning}, pages 1407--1418. PMLR, 2021.

\bibitem{shou2024dynamic}
Yuntao Shou, Tao Meng, Wei Ai, and Keqin Li.
\newblock Dynamic graph neural ordinary differential equation network for multi-modal emotion recognition in conversation.
\newblock {\em arXiv preprint arXiv:2412.02935}, 2024.

\bibitem{yin2024continuous}
Nan Yin, Mengzhu Wan, Li~Shen, Hitesh~Laxmichand Patel, Baopu Li, Bin Gu, and Huan Xiong.
\newblock Continuous spiking graph neural networks.
\newblock {\em arXiv preprint arXiv:2404.01897}, 2024.

\bibitem{xiang2024agc}
Yinfeng Xiang, Chao Li, Shibo He, and Jiming Chen.
\newblock Agc-ode: Adaptive graph controlled neural ode for human mobility prediction.
\newblock {\em IEEE Transactions on Intelligent Transportation Systems}, 2024.

\bibitem{wu2024continuously}
Jiajia Wu and Ling Chen.
\newblock Continuously evolving graph neural controlled differential equations for traffic forecasting.
\newblock {\em arXiv preprint arXiv:2401.14695}, 2024.

\bibitem{wan2024epidemiology}
Guancheng Wan, Zewen Liu, Max~SY Lau, B~Aditya Prakash, and Wei Jin.
\newblock Epidemiology-aware neural ode with continuous disease transmission graph.
\newblock {\em arXiv preprint arXiv:2410.00049}, 2024.

\bibitem{koch2024graph}
James Koch, Pranab~Roy Chowdhury, Heng Wan, Parin Bhaduri, Jim Yoon, Vivek Srikrishnan, and W~Brent Daniel.
\newblock Graph neural ordinary differential equations for coarse-grained socioeconomic dynamics.
\newblock {\em arXiv preprint arXiv:2407.18108}, 2024.

\bibitem{sun2024incorporating}
Yue Sun, Chao Chen, Yuesheng Xu, Sihong Xie, Rick~S Blum, and Parv Venkitasubramaniam.
\newblock Incorporating domain differential equations into graph convolutional networks to lower generalization discrepancy.
\newblock {\em arXiv preprint arXiv:2404.01217}, 2024.

\bibitem{wang2024information}
Ding Wang, Wei Zhou, and Songiln Hu.
\newblock Information diffusion prediction with graph neural ordinary differential equation network.
\newblock In {\em Proceedings of the 32nd ACM International Conference on Multimedia}, pages 9699--9708, 2024.

\bibitem{qin2024learning}
Yifang Qin, Wei Ju, Hongjun Wu, Xiao Luo, and Ming Zhang.
\newblock Learning graph ode for continuous-time sequential recommendation.
\newblock {\em IEEE Transactions on Knowledge and Data Engineering}, 36(7):3224--3236, 2024.

\bibitem{luo2023pgode}
Xiao Luo, Yiyang Gu, Huiyu Jiang, Hang Zhou, Jinsheng Huang, Wei Ju, Zhiping Xiao, Ming Zhang, and Yizhou Sun.
\newblock Pgode: Towards high-quality system dynamics modeling.
\newblock {\em arXiv preprint arXiv:2311.06554}, 2023.

\bibitem{kumawat2024stemfold}
Hemant Kumawat, Biswadeep Chakraborty, and Saibal Mukhopadhyay.
\newblock Stemfold: Stochastic temporal manifold for multi-agent interactions in the presence of hidden agents.
\newblock In {\em 6th Annual Learning for Dynamics \& Control Conference}, pages 1427--1439. PMLR, 2024.

\bibitem{chen2024signed}
Lanlan Chen, Kai Wu, Jian Lou, and Jing Liu.
\newblock Signed graph neural ordinary differential equation for modeling continuous-time dynamics.
\newblock In {\em Proceedings of the AAAI Conference on Artificial Intelligence}, volume~38, pages 8292--8301, 2024.

\bibitem{gravina2024temporal}
Alessio Gravina, Daniele Zambon, Davide Bacciu, and Cesare Alippi.
\newblock Temporal graph odes for irregularly-sampled time series.
\newblock {\em arXiv preprint arXiv:2404.19508}, 2024.

\bibitem{li2024recurrent}
Linzhi Li, Xiaofeng Zhou, Guoliang Hu, Shuai Li, and Dongni Jia.
\newblock A recurrent spatio-temporal graph neural network based on latent time graph for multi-channel time series forecasting.
\newblock {\em IEEE Signal Processing Letters}, 2024.

\bibitem{cheng2024framework}
Haodong Cheng, Yingchi Mao, and Xiao Jia.
\newblock A framework based on physics-informed graph neural ode: for continuous spatial-temporal pandemic prediction.
\newblock {\em Applied Intelligence}, 54(24):12661--12675, 2024.

\bibitem{huang2024causal}
Zijie Huang, Jeehyun Hwang, Junkai Zhang, Jinwoo Baik, Weitong Zhang, Dominik Wodarz, Yizhou Sun, Quanquan Gu, and Wei Wang.
\newblock Causal graph ode: Continuous treatment effect modeling in multi-agent dynamical systems.
\newblock In {\em Proceedings of the ACM Web Conference 2024}, pages 4607--4617, 2024.

\bibitem{han2024brainode}
Kaiqiao Han, Yi~Yang, Zijie Huang, Xuan Kan, Yang Yang, Ying Guo, Lifang He, Liang Zhan, Yizhou Sun, Wei Wang, et~al.
\newblock Brainode: Dynamic brain signal analysis via graph-aided neural ordinary differential equations.
\newblock {\em arXiv preprint arXiv:2405.00077}, 2024.

\bibitem{bhaskar2024inferring}
Dhananjay Bhaskar, Daniel~Sumner Magruder, Matheo Morales, Edward De~Brouwer, Aarthi Venkat, Frederik Wenkel, Guy Wolf, and Smita Krishnaswamy.
\newblock Inferring dynamic regulatory interaction graphs from time series data with perturbations.
\newblock In {\em Learning on Graphs Conference}, pages 22--1. PMLR, 2024.

\bibitem{wen2024second}
Song Wen, Hao Wang, Di~Liu, Qilong Zhangli, and Dimitris Metaxas.
\newblock Second-order graph odes for multi-agent trajectory forecasting.
\newblock In {\em Proceedings of the IEEE/CVF Winter Conference on Applications of Computer Vision}, pages 5101--5110, 2024.

\bibitem{niwa2023coordinet}
Kenta Niwa, Naonori Ueda, Hiroshi Sawada, Akinori Fujino, Shoichiro Takeda, Guoqiang Zhang, and W~Bastiaan Kleijn.
\newblock Coordinet: Constrained dynamics learning for state coordination over graph.
\newblock {\em IEEE Transactions on Signal and Information Processing over Networks}, 9:242--257, 2023.

\bibitem{westny2023evaluation}
Theodor Westny, Joel Oskarsson, Bj{\"o}rn Olofsson, and Erik Frisk.
\newblock Evaluation of differentially constrained motion models for graph-based trajectory prediction.
\newblock In {\em 2023 IEEE Intelligent Vehicles Symposium (IV)}, pages 1--8. IEEE, 2023.

\bibitem{westny2023mtp}
Theodor Westny, Joel Oskarsson, Bj{\"o}rn Olofsson, and Erik Frisk.
\newblock Mtp-go: Graph-based probabilistic multi-agent trajectory prediction with neural odes.
\newblock {\em IEEE Transactions on Intelligent Vehicles}, 8(9):4223--4236, 2023.

\bibitem{luo2023hope}
Xiao Luo, Jingyang Yuan, Zijie Huang, Huiyu Jiang, Yifang Qin, Wei Ju, Ming Zhang, and Yizhou Sun.
\newblock Hope: High-order graph ode for modeling interacting dynamics.
\newblock In {\em International conference on machine learning}, pages 23124--23139. PMLR, 2023.

\bibitem{qin2023learning}
Tiexin Qin, Benjamin Walker, Terry Lyons, Hong Yan, and Haoliang Li.
\newblock Learning dynamic graph embeddings with neural controlled differential equations.
\newblock {\em arXiv preprint arXiv:2302.11354}, 2023.

\bibitem{uwamichi2024integrating}
Masahito Uwamichi, Simon~K. Schnyder, Tetsuya~J. Kobayashi, and Satoshi Sawai.
\newblock Integrating {GNN} and neural {ODE}s for estimating non-reciprocal two-body interactions in mixed-species collective motion.
\newblock In {\em The Thirty-eighth Annual Conference on Neural Information Processing Systems}, 2024.

\bibitem{eliasof2024feature}
Moshe Eliasof, Eldad Haber, and Eran Treister.
\newblock Feature transportation improves graph neural networks.
\newblock In {\em Proceedings of the AAAI Conference on Artificial Intelligence}, volume~38, pages 11874--11882, 2024.

\bibitem{ding2024predicting}
Yanna Ding, Zijie Huang, Malik Magdon-Ismail, and Jianxi Gao.
\newblock Predicting time series of networked dynamical systems without knowing topology.
\newblock {\em arXiv preprint arXiv:2412.18734}, 2024.

\bibitem{xing2023hdg}
Yucheng Xing and Xin Wang.
\newblock Hdg-ode: A hierarchical continuous-time model for human pose forecasting.
\newblock In {\em Proceedings of the IEEE/CVF International Conference on Computer Vision}, pages 14700--14712, 2023.

\bibitem{huang2023tango}
Zijie Huang, Wanjia Zhao, Jingdong Gao, Ziniu Hu, Xiao Luo, Yadi Cao, Yuanzhou Chen, Yizhou Sun, and Wei Wang.
\newblock Tango: Time-reversal latent graphode for multi-agent dynamical systems.
\newblock {\em arXiv preprint arXiv:2310.06427}, 2023.

\bibitem{chu2024adaptive}
Zihao Chu, Wenming Ma, Mingqi Li, and Hao Chen.
\newblock Adaptive decision spatio-temporal neural ode for traffic flow forecasting with multi-kernel temporal dynamic dilation convolution.
\newblock {\em Neural Networks}, 179:106549, 2024.

\bibitem{zhang2024we}
Weizhi Zhang, Liangwei Yang, Zihe Song, Henry~Peng Zou, Ke~Xu, Liancheng Fang, and Philip~S Yu.
\newblock Do we really need graph convolution during training? light post-training graph-ode for efficient recommendation.
\newblock In {\em Proceedings of the 33rd ACM International Conference on Information and Knowledge Management}, pages 3248--3258, 2024.

\bibitem{bishnoi2022enhancing}
Suresh Bishnoi, Ravinder Bhattoo, Sayan Ranu, and NM~Krishnan.
\newblock Enhancing the inductive biases of graph neural ode for modeling dynamical systems.
\newblock {\em arXiv preprint arXiv:2209.10740}, 2022.

\bibitem{bazgir2023integration}
Omid Bazgir, Zichen Wang, Ji~Won Park, Marc Hafner, and James Lu.
\newblock Integration of graph neural network and neural-odes for tumor dynamic prediction.
\newblock {\em arXiv preprint arXiv:2310.00926}, 2023.

\bibitem{mao2024mpstan}
Junkai Mao, Yuexing Han, and Bing Wang.
\newblock Mpstan: Metapopulation-based spatio--temporal attention network for epidemic forecasting.
\newblock {\em Entropy}, 26(4):278, 2024.

\bibitem{qiu2024neural}
Haiquan Qiu, Shuzhi Liu, and Quanming Yao.
\newblock Neural symbolic regression of complex network dynamics.
\newblock {\em arXiv preprint arXiv:2410.11185}, 2024.

\bibitem{mo2024pi}
Zhaobin Mo, Yongjie Fu, and Xuan Di.
\newblock Pi-neugode: Physics-informed graph neural ordinary differential equations for spatiotemporal trajectory prediction.
\newblock In {\em Proceedings of the 23rd International Conference on Autonomous Agents and Multiagent Systems}, pages 1418--1426, 2024.

\bibitem{li2024predicting}
Ruikun Li, Huandong Wang, Jinghua Piao, Qingmin Liao, and Yong Li.
\newblock Predicting long-term dynamics of complex networks via identifying skeleton in hyperbolic space.
\newblock In {\em Proceedings of the 30th ACM SIGKDD Conference on Knowledge Discovery and Data Mining}, pages 1655--1666, 2024.

\bibitem{ma2024spatio}
Wenming Ma, Zihao Chu, Hao Chen, and Mingqi Li.
\newblock Spatio-temporal envolutional graph neural network for traffic flow prediction in uav-based urban traffic monitoring system.
\newblock {\em Scientific Reports}, 14(1):26800, 2024.

\bibitem{wu2024pure}
Hao Wu, Changhu Wang, Fan Xu, Jinbao Xue, Chong Chen, Xian-Sheng Hua, and Xiao Luo.
\newblock Pure: Prompt evolution with graph ode for out-of-distribution fluid dynamics modeling.
\newblock {\em Advances in Neural Information Processing Systems}, 37:104965--104994, 2024.

\bibitem{kosma2023neural}
Chrysoula Kosma, Giannis Nikolentzos, George Panagopoulos, Jean-Marc Steyaert, and Michalis Vazirgiannis.
\newblock Neural ordinary differential equations for modeling epidemic spreading.
\newblock {\em Transactions on Machine Learning Research}, 2023.

\bibitem{gravina2022anti}
Alessio Gravina, Davide Bacciu, and Claudio Gallicchio.
\newblock Anti-symmetric dgn: a stable architecture for deep graph networks.
\newblock {\em arXiv preprint arXiv:2210.09789}, 2022.

\bibitem{zhong2023attention}
Weiheng Zhong, Hadi Meidani, and Jane Macfarlane.
\newblock Attention-based spatial-temporal graph neural ode for traffic prediction.
\newblock {\em arXiv preprint arXiv:2305.00985}, 2023.

\bibitem{jiang2023cf}
Song Jiang, Zijie Huang, Xiao Luo, and Yizhou Sun.
\newblock Cf-gode: Continuous-time causal inference for multi-agent dynamical systems.
\newblock In {\em Proceedings of the 29th ACM SIGKDD Conference on Knowledge Discovery and Data Mining}, pages 997--1009, 2023.

\bibitem{xiong2023diffusion}
Ni~Xiong, Yan Yang, Yongquan Jiang, and Xiaocao Ouyang.
\newblock Diffusion graph neural ordinary differential equation network for traffic prediction.
\newblock In {\em 2023 International Joint Conference on Neural Networks (IJCNN)}, pages 1--8. IEEE, 2023.

\bibitem{choi2023gread}
Jeongwhan Choi, Seoyoung Hong, Noseong Park, and Sung-Bae Cho.
\newblock Gread: Graph neural reaction-diffusion networks.
\newblock In {\em International Conference on Machine Learning}, pages 5722--5747. PMLR, 2023.

\bibitem{liu2023graph}
Zibo Liu, Parshin Shojaee, and Chandan~K Reddy.
\newblock Graph-based multi-ode neural networks for spatio-temporal traffic forecasting.
\newblock {\em arXiv preprint arXiv:2305.18687}, 2023.

\bibitem{zhan2023learning}
Fei Zhan, Xiaofeng Zhou, Shuai Li, Dongni Jia, and Hong Song.
\newblock Learning latent odes with graph rnn for multi-channel time series forecasting.
\newblock {\em IEEE Signal Processing Letters}, 30:1432--1436, 2023.

\bibitem{zeng2023long}
Hui Zeng, Chaojie Jiang, Yuanchun Lan, Xiaohui Huang, Junyang Wang, and Xinhua Yuan.
\newblock Long short-term fusion spatial-temporal graph convolutional networks for traffic flow forecasting.
\newblock {\em Electronics}, 12(1):238, 2023.

\bibitem{sun2023reaction}
Yue Sun, Chao Chen, Yuesheng Xu, Sihong Xie, Rick~S Blum, and Parv Venkitasubramaniam.
\newblock Reaction-diffusion graph ordinary differential equation networks: Traffic-law-informed speed prediction under mismatched data.
\newblock The 12th International Workshop on Urban Computing, held in conjunction with~…, 2023.

\bibitem{wang2023temporal}
Shuang Wang, Hong Dai, Lin Bai, Chengrui Liu, and Junhong Chen.
\newblock Temporal branching-graph neural ode without prior structure for traffic flow forecasting.
\newblock {\em Eng. Let}, 31:1534, 2023.

\bibitem{oskarsson2023temporal}
Joel Oskarsson, Per Sid{\'e}n, and Fredrik Lindsten.
\newblock Temporal graph neural networks for irregular data.
\newblock In {\em International Conference on Artificial Intelligence and Statistics}, pages 4515--4531. PMLR, 2023.

\bibitem{xie2023temporal}
Yi~Xie, Yun Xiong, Jiawei Zhang, Chao Chen, Yao Zhang, Jie Zhao, Yizhu Jiao, Jinjing Zhao, and Yangyong Zhu.
\newblock Temporal super-resolution traffic flow forecasting via continuous-time network dynamics.
\newblock {\em Knowledge and Information Systems}, 65(11):4687--4712, 2023.

\bibitem{asikis2022neural}
Thomas Asikis, Lucas B{\"o}ttcher, and Nino Antulov-Fantulin.
\newblock Neural ordinary differential equation control of dynamics on graphs.
\newblock {\em Physical Review Research}, 4(1):013221, 2022.

\bibitem{wen2022social}
Song Wen, Hao Wang, and Dimitris Metaxas.
\newblock Social ode: Multi-agent trajectory forecasting with neural ordinary differential equations.
\newblock In {\em European Conference on Computer Vision}, pages 217--233. Springer, 2022.

\bibitem{fang2021spatial}
Zheng Fang, Qingqing Long, Guojie Song, and Kunqing Xie.
\newblock Spatial-temporal graph ode networks for traffic flow forecasting.
\newblock In {\em Proceedings of the 27th ACM SIGKDD conference on knowledge discovery \& data mining}, pages 364--373, 2021.

\bibitem{desai2021variational}
Shaan~A Desai, Marios Mattheakis, and Stephen~J Roberts.
\newblock Variational integrator graph networks for learning energy-conserving dynamical systems.
\newblock {\em Physical Review E}, 104(3):035310, 2021.

\bibitem{mercatali2024graph}
Giangiacomo Mercatali, Andre Freitas, and Jie Chen.
\newblock Graph neural flows for unveiling systemic interactions among irregularly sampled time series.
\newblock {\em arXiv preprint arXiv:2410.14030}, 2024.

\bibitem{liu2023segno}
Yang Liu, Jiashun Cheng, Haihong Zhao, Tingyang Xu, Peilin Zhao, Fugee Tsung, Jia Li, and Yu~Rong.
\newblock Segno: Generalizing equivariant graph neural networks with physical inductive biases.
\newblock {\em arXiv preprint arXiv:2308.13212}, 2023.

\bibitem{thangamuthu2022unravelling}
Abishek Thangamuthu, Gunjan Kumar, Suresh Bishnoi, Ravinder Bhattoo, NM~Krishnan, and Sayan Ranu.
\newblock Unravelling the performance of physics-informed graph neural networks for dynamical systems.
\newblock {\em Advances in Neural Information Processing Systems}, 35:3691--3702, 2022.

\bibitem{ding2024architecture}
Yanna Ding, Zijie Huang, Xiao Shou, Yihang Guo, Yizhou Sun, and Jianxi Gao.
\newblock Architecture-aware learning curve extrapolation via graph ordinary differential equation.
\newblock {\em arXiv preprint arXiv:2412.15554}, 2024.

\bibitem{sun2024graph}
Fang Sun, Zijie Huang, Haixin Wang, Yadi Cao, Xiao Luo, Wei Wang, and Yizhou Sun.
\newblock Graph fourier neural odes: Bridging spatial and temporal multiscales in molecular dynamics.
\newblock {\em arXiv preprint arXiv:2411.01600}, 2024.

\bibitem{cui2024predictive}
Zhe Cui, Di~Zang, Hong Zhu, and Keshuang Tang.
\newblock Predictive and multigranularity resilience assessment of urban transportation based on neural controlled differential equation.
\newblock {\em IEEE Transactions on Reliability}, 2024.

\bibitem{choi2022graph}
Jeongwhan Choi, Hwangyong Choi, Jeehyun Hwang, and Noseong Park.
\newblock Graph neural controlled differential equations for traffic forecasting.
\newblock In {\em Proceedings of the AAAI conference on artificial intelligence}, volume~36, pages 6367--6374, 2022.

\bibitem{bergna2024uncertainty}
Richard Bergna, Sergio Calvo-Ordonez, Felix~L Opolka, Pietro Li{\`o}, and Jose~Miguel Hernandez-Lobato.
\newblock Uncertainty modeling in graph neural networks via stochastic differential equations.
\newblock {\em arXiv preprint arXiv:2408.16115}, 2024.

\bibitem{cao_inductive_2021}
Huafeng Cao, Zhongbao Zhang, Li~Sun, and Zhi Wang.
\newblock Inductive and irregular dynamic network representation based on ordinary differential equations.
\newblock {\em Knowledge-Based Systems}, 227:107271, September 2021.

\bibitem{jin2022neural}
Ming Jin, Yuan-Fang Li, and Shirui Pan.
\newblock Neural temporal walks: Motif-aware representation learning on continuous-time dynamic graphs.
\newblock {\em Advances in Neural Information Processing Systems}, 35:19874--19886, 2022.

\bibitem{choi_lt-ocf_2021}
Jeongwhan Choi, Jinsung Jeon, and Noseong Park.
\newblock {LT}-{OCF}: {Learnable}-{Time} {ODE}-based {Collaborative} {Filtering}, August 2021.
\newblock arXiv:2108.06208 [cs].

\bibitem{wang2024contig}
Zihui Wang, Peizhen Yang, Xiaoliang Fan, Xu~Yan, Zonghan Wu, Shirui Pan, Longbiao Chen, Yu~Zang, Cheng Wang, and Rongshan Yu.
\newblock Contig: Continuous representation learning on temporal interaction graphs.
\newblock {\em Neural Networks}, 172:106151, 2024.

\end{thebibliography}
\bibliographystyle{unsrt}
\clearpage
\onecolumn
\begin{appendices}
\section{Summary of Graph NDEs}
\label{appendix: A}

\begin{table*}[h]
\caption{
\textbf{Comprehensive Summary of Graph Neural Differential Equations}.
}
\vskip -1em
\centering
{
\resizebox{\linewidth}{!}{
\setlength\tabcolsep{10pt}
\renewcommand\arraystretch{1.2}
\begin{tabular}{@{}c|c|c|c|c@{}} 
\hline
\thickhline

\textbf{Task}   & \textbf{Paper} & \textbf{Role of GNN} & \textbf{DE Type} & \textbf{Graph Construction} \\ 
\hline

\cline{1-5}   
\multirow{9}{*}{\textbf{Node/Graph Classification}} 
  & ~\cite{poli2019graph, huang2024embedding} 
       & \multirow{1}{*}{DE} 
       & \multirow{1}{*}{First Order ODE} 
       & \multirow{1}{*}{Dynamic} \\
\cline{2-5}
  & ~\cite{zang_neural_2020, xhonneux2020continuous, thorpe_grand_2022, maskey_fractional_nodate, zhang2024unleashing, zhao2024adversarial, cui2024graph, wang_acmp_2023, chamberlain2021grand, shou2024dynamic}
       & \multirow{1}{*}{DE} 
       & \multirow{1}{*}{First Order ODE} 
       & \multirow{1}{*}{Static} \\
\cline{2-5}
  & ~\cite{rusch_graph-coupled_2022}
       & DE 
       & Second Order ODE 
       & Static \\
\cline{2-5}
  & ~\cite{kang2024coupling}
       & DE 
       & Fractional Order ODE 
       & Static \\
\cline{2-5}
  & ~\cite{bergna_graph_2023}
       & Encoder 
       & First Order SDE 
       & Static \\
\cline{2-5}
  & ~\cite{bergna_uncertainty_2024, lin2024graph}
       & \multirow{1}{*}{DE}
       & \multirow{1}{*}{First Order SDE}
       & \multirow{1}{*}{Static} \\
\cline{2-5}
  & ~\cite{yan_hypergraph_2024}
       & DE
       & First Order ODE
       & Static \\
\cline{2-5}
  & ~\cite{zhao2023graph, song2022robustness, chamberlain_beltrami_2021}
       & \multirow{1}{*}{DE}
       & \multirow{1}{*}{First Order PDE}
       & \multirow{1}{*}{Static} \\
\cline{2-5}
  & ~\cite{yin2024continuous}
       & \multirow{1}{*}{DE}
       & \multirow{1}{*}{First Order PDE \& Second Order PDE}
       & \multirow{1}{*}{Static} \\


\hline

\multirow{21}{*}{\textbf{Forecasting}}
  & ~\cite{xing_aggdn_2024}
       & DE
       & First Order ODE \& First Order SDE
       & Dynamic \\
\cline{2-5}
  & ~\cite{li_physics-informed_2024, jin_multivariate_2023, xiang2024agc, wu2024continuously, wan2024epidemiology, koch2024graph, sun2024incorporating, wang2024information, qin2024learning, luo2023pgode, kumawat2024stemfold, chen2024signed, gravina2024temporal}
       & \multirow{1}{*}{DE}
       & \multirow{1}{*}{First Order ODE}
       & \multirow{1}{*}{Dynamic} \\
\cline{2-5}
  & ~\cite{Huang2021CGODE, li2024recurrent, cheng2024framework, huang2024causal}
       & Encoder \& DE
       & First Order ODE
       & Dynamic \\
\cline{2-5}
  & ~\cite{han2024brainode, bhaskar2024inferring, wang_causalgnn_2022}
       & Encoder
       & First Order ODE
       & Dynamic \\
\cline{2-5}
  & ~\cite{wen2024second, niwa2023coordinet, westny2023evaluation}
       & DE
       & Second Order ODE
       & Dynamic \\ 
\cline{2-5}
  & ~\cite{westny2023mtp}
       & Encoder
       & Second Order ODE
       & Dynamic \\ 
\cline{2-5}
  & ~\cite{luo2023hope}
       & Encoder \& DE
       & Second Order ODE
       & Dynamic \\ 
\cline{2-5}
  & ~\cite{qin2023learning}
       & DE
       & First-order NCDE (Neural Controlled Differential Equation)
       & Dynamic \\
\cline{2-5}
  & ~\cite{liang2024dynamic}
       & Encoder
       & First Order SDE
       & Dynamic \\
\cline{2-5}
  & ~\cite{cranmer2020lagrangian}
       & DE
       & Second Order PDE
       & Dynamic \\
\cline{2-5}
  & ~\cite{uwamichi2024integrating}
       & DE
       & First Order ODE \& First Order SDE
       & Static \\
\cline{2-5}
  & ~\cite{kumar_grade_2021}
       & DE
       & Second Order PDE
       & Static \\
\cline{2-5}
  & ~\cite{bryutkin_hamlet_2024}
       & Encoder
       & First Order PDE
       & Static \\
\cline{2-5}
  & ~\cite{iakovlev_learning_2021}
       & DE
       & High Order PDE
       & Static \\
\cline{2-5}
  & ~\cite{long2024unveiling, eliasof2024feature}
       & DE
       & First Order DDE
       & Static \\
\cline{2-5}
  & ~\cite{Huang2023GGODE, Huang2020LGODE, ding2024predicting, xing2023hdg, huang2023tango, huang2020learning}
       & \multirow{1}{*}{Encoder \& DE}
       & \multirow{1}{*}{First Order ODE}
       & \multirow{1}{*}{Static} \\
\cline{2-5}
  & ~\cite{yao_spatio-temporal_2023, chu2024adaptive, zhang2024we, bishnoi2022enhancing, bazgir2023integration, mao2024mpstan, qiu2024neural, mo2024pi, li2024predicting, ma2024spatio, shi2024towards, wu2024pure, yuan2024egode, kosma2023neural, gravina2022anti, zhong2023attention, jiang2023cf, xiong2023diffusion, choi2023gread, liu2023graph, zhan2023learning, zeng2023long, sun2023reaction, wang2023temporal, oskarsson2023temporal, xie2023temporal, asikis2022neural, wen2022social, huang2021str, fang2021spatial, desai2021variational, mercatali2024graph}
       & DE
       & First Order ODE
       & Static \\
\cline{2-5}
  & ~\cite{liu2023segno, thangamuthu2022unravelling}
       & DE
       & Second Order ODE
       & Static \\
\cline{2-5}
  & ~\cite{ding2024architecture, sun2024graph}
       & Encoder
       & First Order ODE
       & Static \\
\cline{2-5}
  & ~\cite{cui2024predictive, choi2022graph}
       & DE
       & First-order NCDE (Neural Controlled Differential Equation)
       & Static \\
\cline{2-5}
  & ~\cite{bishnoi2024brognet, bergna2024uncertainty}
       & DE
       & First Order SDE
       & Static \\
\cline{2-5}
  & ~\cite{sanchez2019hamiltonian}
       & DE
       & First Order PDE
       & Static \\
\cline{2-5}
  & ~\cite{xiong2023diffusion}
       & DE \& Decoder
       & First Order ODE
       & Static \\
       
\hline
\multirow{2}{*}{\textbf{Link Prediction}}
  & ~\cite{cao_inductive_2021, jin2022neural}
       & \multirow{1}{*}{DE}
       & \multirow{1}{*}{First Order ODE}
       & \multirow{1}{*}{Dynamic} \\
\cline{2-5}
  & ~\cite{xunode}
       & \multirow{1}{*}{Decoder}
       & \multirow{1}{*}{First Order ODE}
       & \multirow{1}{*}{Dynamic} \\

\hline

\multirow{3}{*}{\textbf{Ranking}}
  & ~\cite{choi_lt-ocf_2021}
       & DE
       & First Order ODE
       & Static \\
\cline{2-5}
  & ~\cite{zhang_cope_2021}
       & DE
       & First Order ODE
       & Dynamic \\
\cline{2-5}
  & ~\cite{wang2024contig}
       & Decoder
       & First Order ODE
       & Dynamic \\

\hline

\multirow{4}{*}{\textbf{Graph Generation}}
  & ~\cite{verma_modular_nodate}
       & \multirow{1}{*}{DE}
       & \multirow{1}{*}{First Order PDE}
       & \multirow{1}{*}{Dynamic} \\
\cline{2-5}
  & ~\cite{niu2020permutation, huang_graphgdp_2022, jo2022score}
       & \multirow{1}{*}{DE}
       & \multirow{1}{*}{First Order SDE}
       & \multirow{1}{*}{Dynamic} \\
\cline{2-5}
  & ~\cite{huang2023conditional}
       & Encoder
       & Third Order SDE
       & Dynamic \\
\hline

\end{tabular}}}
\end{table*}

\end{appendices}

\end{document}